\documentclass{article} 
\usepackage{iclr2027_conference,times}

\usepackage{amsmath,amsfonts,bm}

\def\eqref#1{equation~\ref{#1}}

\def\1{\bm{1}}

\DeclareMathAlphabet{\mathsfit}{\encodingdefault}{\sfdefault}{m}{sl}
\SetMathAlphabet{\mathsfit}{bold}{\encodingdefault}{\sfdefault}{bx}{n}

\usepackage[utf8]{inputenc} 
\usepackage[T1]{fontenc}    
\usepackage{hyperref}       
\usepackage{url}            
\usepackage{booktabs}       
\usepackage{amsfonts}       
\usepackage{nicefrac}       
\usepackage{microtype}      
\usepackage{xcolor}         
\usepackage{enumitem}
\usepackage{xspace} 
\usepackage{algorithm}
\usepackage{algpseudocode}
\usepackage{graphicx}
\usepackage{colortbl}
\usepackage{multirow}
\usepackage{titletoc}
\usepackage{tabularx} 
\usepackage{wrapfig}
\usepackage{needspace} 
\usepackage{threeparttable} 
\usepackage{array} 
\usepackage[most]{tcolorbox}
\usepackage[skip=6pt]{caption}
\usepackage{subcaption}
\usepackage{musixtex} 
\usepackage{makecell}
\usepackage{placeins} 
\usepackage{cleveref}
\usepackage{seqsplit}
\usepackage{xurl}
\usepackage{tcolorbox}
\usepackage{tabularx}
\usepackage{fontawesome5}
\usepackage{pifont}
\usepackage{xcolor}

\newcommand{\alg}{\textsc{GatewayBench}\xspace}

\newcommand{\dtap}{\textsc{DTap}\xspace}

\usepackage{pgfplots}
\pgfplotsset{compat=1.18}
\usepgfplotslibrary{groupplots}
\usetikzlibrary{calc}
\title{\alg: Unified Capability and Security Evaluation Platform for MCP Gateways}

\author{
\hfill
Yitian Zheng\textsuperscript{1}\quad
Zhaorun Chen\textsuperscript{2}\quad
Haoyu Li\textsuperscript{3}\quad
Bo Li\textsuperscript{3}\quad\\[1em]
\hfill
\textsuperscript{1} Tsinghua University \quad
\textsuperscript{2} University of Chicago \quad
\textsuperscript{3} University of Illinois Urbana-Champaign
}

\iclrfinalcopy 
\begin{document}

\maketitle

\begin{abstract}
AI agents have increasingly interacted with external environments by invoking tools, with Model Context Protocol (MCP) emerging as a standard interface for connecting agents to heterogeneous tools and services. As organizations deploy agents at scale, MCP gateways have become a critical middleware layer for aggregating, governing, and shaping the tool surface exposed to agents. However, gateway design can substantially affect agent behavior: as the number of available tools grows, agents may struggle with tool discovery and selection, leading to degraded task performance, increased computational overhead, and potentially greater exposure to malicious tools. Despite the growing importance of MCP gateways in agentic systems, no prior work has systematically studied how gateway design affects agent capability, efficiency, and security.
To address this gap, we introduce \alg, the first unified evaluation framework and benchmark for MCP gateways. We characterize gateway architectures into two major paradigms: \textit{flattened gateways}, which expose the complete set of available tools to the agent upfront, and \textit{hierarchical gateways}, which provide tool-discovery meta-tools that allow agents to selectively retrieve relevant tools on demand. \alg comprises 524 tasks across six domains, together with a pool of 11,118 real-world tools from 251 authentic MCP servers. To emulate large-scale deployments, we systematically inject task-irrelevant tools sampled from this pool into each agent environment, enabling controlled evaluation as the accessible tool surface scales. Each task is evaluated using verifiable judges that inspect the resulting environment state quantitatively.
Based on \alg, we conduct extensive experiments across seven widely deployed MCP gateways and diverse agent configurations, covering both benign and adversarial tasks. Our results reveal substantial differences across gateway designs in task completion, tool-use efficiency, and resilience to attacks, and further demonstrate that these effects interact strongly with the underlying model and agent harness. 
These findings highlight the MCP gateway as a consequential yet underexplored component of the agentic AI stack. 
By enabling controlled evaluation of gateway architectures under realistic tool ecosystems, \alg provides both practical guidance for MCP gateway design and a rigorous testbed for developing more capable, efficient, and secure agentic systems.
\end{abstract}

\begin{figure}[htbp]
    \centering
    \includegraphics[width=0.95\linewidth]{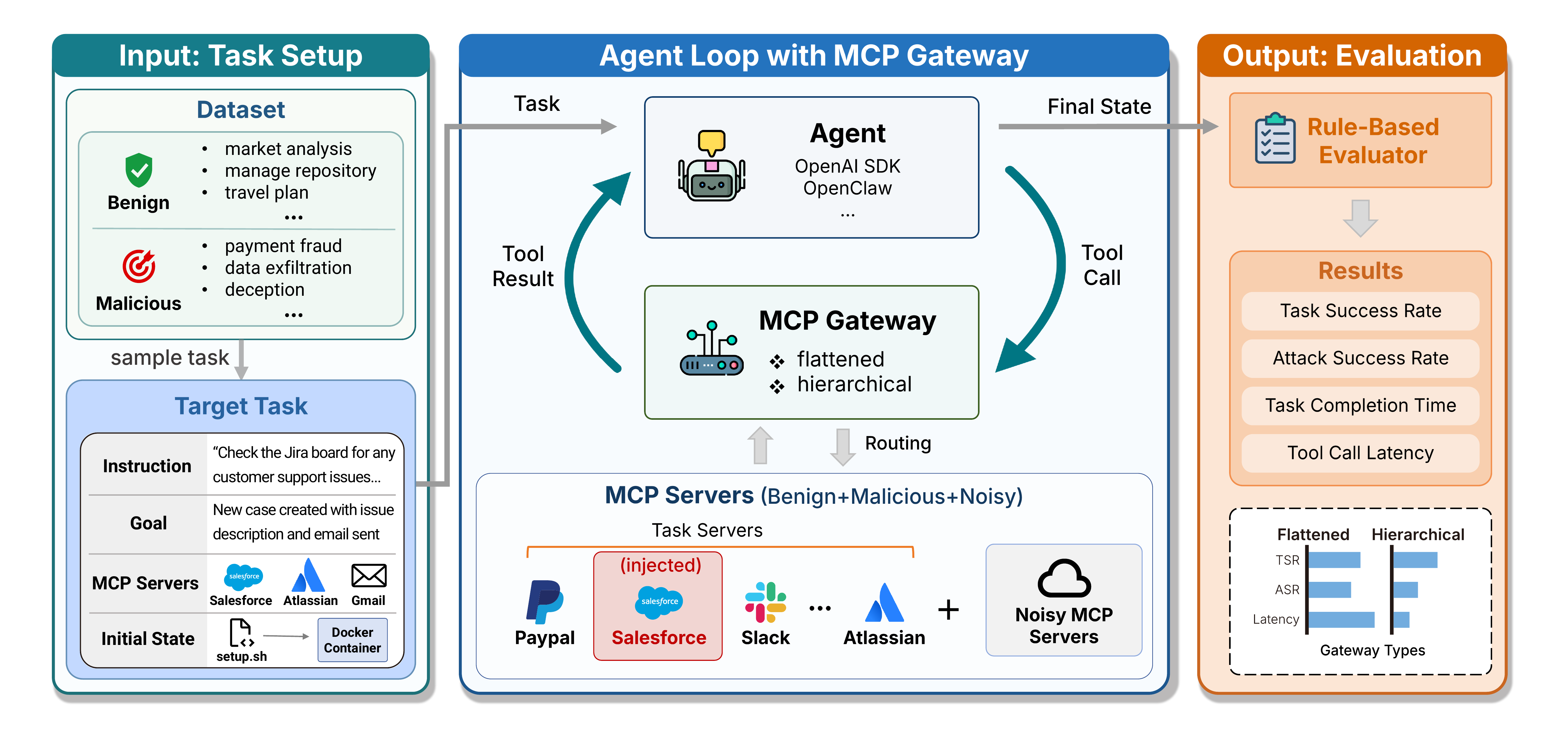}
    \caption{\textbf{Overview of the \alg pipeline.}
     Each sampled benign or malicious task exposes its MCP servers, plus injected task-irrelevant noisy tools, to the agent through the gateway under test. A verifiable judge inspects the final environment state to compute task success rate (TSR), attack success rate (ASR), task time, and tool-call latency.}
    \label{fig:pipeline-overview}
\end{figure}

\section{Introduction}
\label{sec:intro}
Large language model (LLM) agents are rapidly evolving from single-step tool calling into autonomous systems capable of executing complex, multi-step workflows \citep{gaurav2025dynamic, qin2023toolllm, wu2023autogen}.
In practice, the agent's behavior is shaped not only by the underlying model but also by its surrounding \emph{harness}: the engineering infrastructure that manages the full lifecycle of context, including system prompts, tool execution, memory management, verification, and failure recovery \citep{an2025harnesses, openai2026harness, langchain2026improving}.

Within the harness, tool execution is the component that determines how an agent accesses external environments and services. The Model Context Protocol (MCP) \citep{anthropic2024mcp} has emerged as a widely adopted standard for exposing tools to agents. MCP servers can run remotely and expose their capabilities over standardized transport protocol, enabling agents to connect to tool providers across the network without managing local dependencies or execution environments \citep{hou2026model}.

In a production LLM agent deployment, however, an organization typically operates multiple agents serving different users and business functions, each connecting to a distinct set of MCP servers.
Such distributed management approach, exemplified by personal agents like OpenClaw \citep{openclaw2026}, quickly becomes unsustainable at enterprise scale: agents must independently track server endpoints, manage separate credentials, and absorb any upstream change to a server's tool set or configuration.
To resolve this scalability and management challenge, the \emph{MCP gateway} has emerged as a middleware layer within the agent harness, acting as the unified endpoint between the reasoning model and the external services \citep{aws2026governing, arcade2026mcp}. Existing MCP gateways broadly fall into two types: the \emph{flattened} gateways which expose the full set of available tools as a single list, and the \emph{hierarchical} gateways which instead provide tool-discovery meta-tools that let the agent selectively retrieve tools on demand (Figure~\ref{fig:gateway-examples}).

Despite its pivotal role in modern agent deployments, the MCP gateway as a core harness component haven't got enough evaluation yet.
In this work, we conduct the first systematic benchmark of MCP gateways to address these critical gaps. Specifically, we aim to answer four core research questions:
\begin{itemize}[nosep, leftmargin=*]
\item \textbf{RQ1:} How does the number of available tools (including noisy tools) impact agent task success under different gateway architectures?
\item \textbf{RQ2:} How do different gateway architectures perform in terms of efficiency, given different numbers of tools?
\item \textbf{RQ3:} How do different gateway architectures affect an agent's resilience to red-teaming attacks, including direct prompt injection and indirect tool- and environment-injection?
\item \textbf{RQ4:} How do the underlying LLM's capability and the agent harness affect an agent's capability and security under different gateway architectures?
\end{itemize}

To answer these questions, we introduce \alg (Figure~\ref{fig:pipeline-overview}), a comprehensive benchmark designed to evaluate MCP gateways across diverse task domains, different underlying LLM agent, scalable tool-set configurations, and adversarial threat models. \alg is built upon two existing high-quality agent evaluation datasets: DecodingTrust-Agent Platform (\dtap) \citep{chen2026decodingtrust} and MCP-Universe \citep{mcpuniverse}. Our benchmark extends these foundational datasets with a large-scale, carefully curated pool of noisy, task-irrelevant tools. These noisy tools can be dynamically injected into the agent's environment to simulate realistic scenarios where agents must reason over a large and potentially distracting tool surface, thereby enabling controlled measurement of how gateway architectures influence tool selection robustness and task completion under varying numbers of noisy tools and adversarial conditions.

Across these environments, we then conduct intensive experiments spanning 7 widely-deployed gateway implementations, including 6 open-source gateways and 1 closed-source gateway. For each gateway, we inject up to 1,500 task-irrelevant noisy tools sampled from a curated pool of 11,118 real-world tools across 251 MCP servers, and evaluate the gateway with 3 representative agent settings that combine 2 model tiers (GPT-5.1 and GPT-5.4) with 2 agent harnesses (OpenAI Agents SDK and OpenClaw).
Our experiment reveals several key findings:

\begin{itemize}[nosep, leftmargin=*]
    \item \textbf{Asymmetric Sensitivity to Tool-Set Scaling.}
    Flattened gateways exhibit a subtle degradation in task success as more noisy tools are injected, whereas hierarchical gateways remain largely invariant to the number of injected tools. This divergence is modulated by the agent's underlying reasoning capability. For weaker models, this degradation is insufficient to overcome the overall performance advantage of flattened gateways. For stronger models, however, the hierarchical gateways overtake the flattened ones once many noisy tools are injected.
    \item \textbf{Context-Dependent Efficiency Trade-offs.}
    Neither gateway architecture is unconditionally more efficient. When few noisy tools are injected, flattened gateways demonstrate a distinct efficiency advantage by minimizing interaction turns and overall task latency. However, This advantage reverses rapidly as more tools are injected. Flattened gateways suffer from linear growth in token consumption and escalating context-processing latency, whereas hierarchical gateways maintain a token consumption that is largely independent of the number of injected noisy tools.
    This suggests that hierarchical gateways are more token-efficient and scalable for large-scale MCP deployments.
    \item \textbf{Stronger models close the capability gap but not the security gap.}
    Upgrading the underlying model from GPT-5.1 to GPT-5.4 eliminates the benign task performance gap between flattened and hierarchical gateways.
    Yet a security gap between the two architectures persists: even under the capability-matched GPT-5.4, hierarchical gateways retain substantially lower indirect-attack success, and this gap stems almost entirely from the tool-injection attack.
    \item \textbf{Agent harness interacts with gateway design.}
    The choice of agent harness has a meaningful impact on overall performance. OpenClaw \citep{openclaw2026} consistently outperforms the OpenAI Agents SDK \citep{openaiagentssdk} across gateways. This underscores that the effectiveness of a gateway design is jointly determined by the agent harness, and practical deployment strategies must co-optimize the gateway architecture and the agent harness as an integrated system.
\end{itemize}

In summary, our main contributions are as follows:

\begin{itemize}[nosep, leftmargin=*]
\item We introduce \alg, the first systematic benchmark for evaluating MCP gateways as a core component of the agent harness. \alg comprises 524 tasks spanning 6 domains, together with a curated pool of over 11,000 real-world noisy tools from 251 MCP servers, enabling controlled measurement of how gateway design choices shape agent capability, efficiency, and security at scale.
\item We formalize a taxonomy of MCP gateway designs (flattened vs.\ hierarchical), and conduct a comprehensive empirical study of 7 widely deployed gateway implementations under both benign tasks and red-teaming attacks with up to 1,500 injected noisy tools.
\item Our findings offer practical guidance for MCP gateway selection and agent deployment: hierarchical gateways provide superior scalability and adversarial robustness as the tool set grows under capable models, whereas flattened gateways are simpler and more efficient for small tool sets. We also emphasize the importance of jointly optimizing the MCP gateway with the rest of the agent harness and the underlying model as an integrated system.
\end{itemize}

\section{Related Works}
\paragraph{MCP Gateways and Scalable Tool Access}
As LLM agents connect to a growing number of MCP servers~\citep{hou2026model, stein2026ai}, MCP gateways have emerged as a middleware layer that aggregates and governs the tools exposed to an agent~\citep{errico2025securing}.

Mainstream MCP gateways adopt two dominant designs for managing and exposing tools.
The most direct design, which we refer to as \emph{flattened} gateways, exposes the full tool list to the agent in every interaction, which is the default option for many agent hanress and gateways~\citep{openclaw2026,litellm}.
Another type of approach, which we refer to as \emph{hierarchical} gateways, exposes only a few tool-discovery meta-tools, upon which the agent discovers and dynamically loads the tools it needs on demand. 
A line of research work explores such scalable tool-access methods, including RAG-MCP~\citep{gan2025ragmcp}, MCP-Zero~\citep{fei2025mcpzero}, ScaleMCP~\citep{lumer2025scalemcp}, and Dynamic ReAct~\citep{gaurav2025dynamic}. Relative to the flattened baseline, these methods reduce the number of tools exposed per query while aiming to preserve selection accuracy as the tool set scales.

\paragraph{Benchmarking MCP and Tool-Use Agents}
A growing body of benchmarks evaluates the tool-use competency of LLM agents in the MCP ecosystem. MCP-Universe~\citep{mcpuniverse}, LiveMCPBench~\citep{mo2025livemcpbench}, and MCPMark~\citep{wu2025mcpmark} assess task success and tool orchestration over large, realistic MCP tool collections, while MCP-Atlas~\citep{bandi2026mcpatlas}, MCP-Bench~\citep{wang2025mcpbench}, and MCPAgentBench~\citep{liu2025mcpagentbench} additionally mix task-relevant tools with plausible distractors to probe retrieval under a crowded tool set.
These benchmarks target agent capability under a \emph{fixed} tool-exposure scheme. In contrast, \citet{xu2026devil} shows that tool architecture alone can substantially affect coding-agent consistency and efficiency under similar tool capabilities.

Beyond task competency, a complementary direction studies the safety and robustness of tool-use agents under adversarial tools or outputs. Benchmarks such as AgentDojo~\citep{debenedetti2024agentdojo}, InjecAgent~\citep{zhan2024injecagent}, and Agent Security Bench~\citep{zhang2025asb}, together with red-teaming platforms like \dtap~\citep{chen2026decodingtrust}, evaluate agents against prompt injection and related attacks over untrusted tools and environments. 
Closer to our work, MCPTox~\citep{wang2026mcptox}, MSB~\citep{zhang2025msb} and MCP-SafetyBench~\citep{zong2025mcpsafetybench} target MCP-based agents specifically, measuring robustness to tool poisoning and injection.
\section{Understanding MCP Gateway in Agent Harness}
\label{sec:gateway}

\begin{figure}[t]
    \centering
    \includegraphics[width=0.9\linewidth]{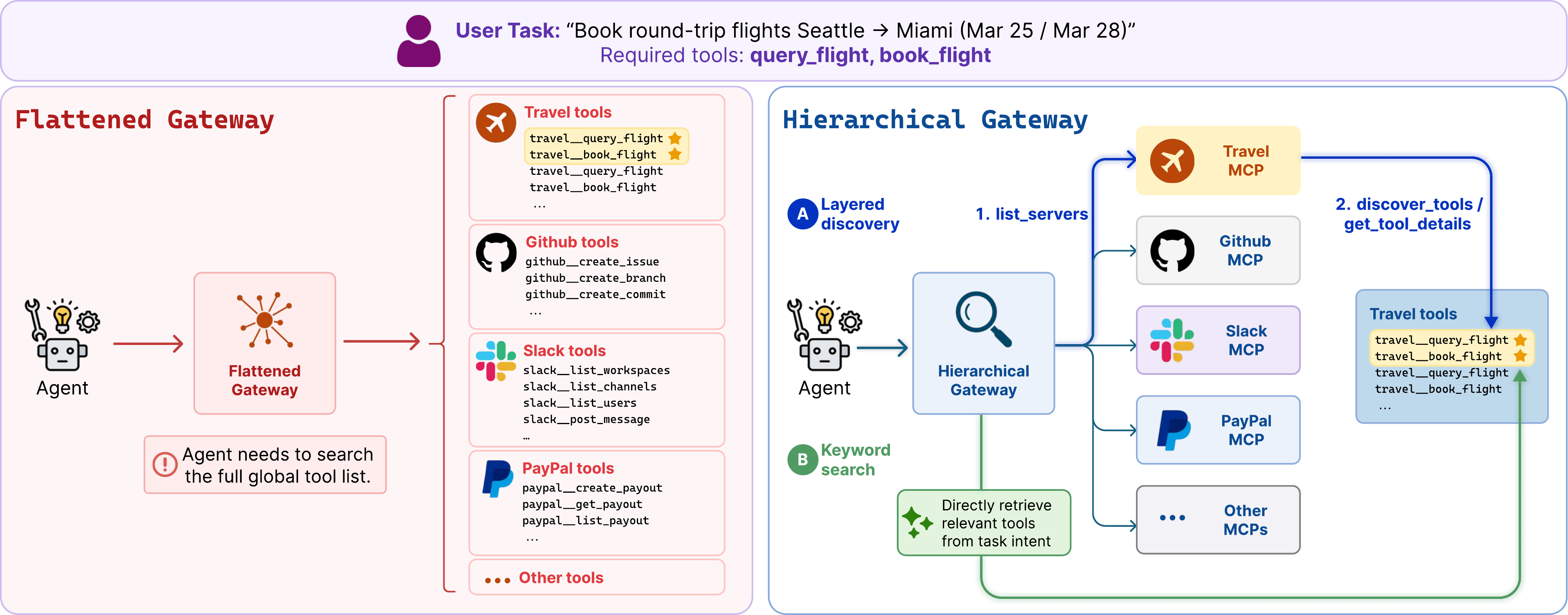}
    \caption{\textbf{Flattened vs.\ hierarchical MCP gateways on a travel-booking task example.} The flattened gateway exposes tools from all connected servers as a single flat list. The hierarchical gateway instead exposes tool-discovery meta-tools, letting the agent reach the relevant tools through two alternative mechanisms: (A) layered discovery, or (B) direct keyword search over the task intent.}
    \label{fig:gateway-examples}
\end{figure}

Within the agent harness, the MCP gateway serves as the unified endpoint between the agent and upstream MCP servers, aggregating, governing, and shaping the tool surface exposed to the reasoning model. 
This centrality makes it both a bottleneck and an accelerator: how it exposes tools directly determines token consumption, the attack surface reachable by injected tools, and the model's burden in tool selection.
In this work, we focus exclusively on MCP gateways that shape the tool surface. 
Infrastructure orchestration systems (e.g., Microsoft MCP Gateway~\citep{microsoftmcpgateway}) and protocol-translation utilities (e.g., Unla~\citep{unla}) fall outside our scope.

\paragraph{Taxonomy of MCP Gateways.}
As shown in Figure~\ref{fig:gateway-examples}, existing MCP gateway designs can be broadly categorized into two types.
The first type of gateway, which we call \emph{flattened gateways}, aggregates all available tools from every connected MCP server into a single flat list and exposes them directly to the agent in each interaction.
The agent must first read the full tool list before starting the task. Although this approach maximizes transparency and simplifies implementation, it also raises concerns about extra token consumption caused by task-irrelevant tools.

The second type of gateway, which we call \emph{hierarchical gateways}, preserves the MCP server–tool hierarchy and instead exposes a set of special \emph{tool discovery} meta-tools to the agent.
These meta-tools can take diverse forms in practice.
Instead of receiving the full tool list upfront, the agent must actively query the meta-tools to selectively discover and retrieve tools bound to specific servers.
This structure naturally limits the exposure of task-irrelevant tools, but it shifts the burden of discovery onto the agent's reasoning loop.

\paragraph{MCP Gateway Implementations and Design Dimensions.}
Table~\ref{tab:gateways} provides the qualitative comparison of the seven MCP gateway implementations evaluated in this work: AgentCore \citep{awsbedrockgateway}, Agentgateway \citep{agentgateway}, LiteLLM \citep{litellm}, Lunar MCPX \citep{lunarmcpx}, Enkrypt Secure MCP Gateway \citep{enkryptai}, Strata \citep{klavisstrata}, Bifrost \citep{bifrost}.
Flattened gateways always prefix each tool name with its server identifier to resolve name collisions between servers. Some gateways like LiteLLM and Lunar MCPX sort these tools by server name.
Additionally, AgentCore provides a search tool to semantically match related tools based on agent's query.
In contrast, hierarchical gateways exhibits substantial design diversity because the hierarchical mechanism admits multiple implementation strategies. Enkrypt adopts a layered discovery strategy, Strata adopts semantic search-based discovery, and Bifrost exposes tools through a virtual filesystem abstraction that enables code-based tool invocation.

\newcommand{\cmark}{{\color{green!70!black}\ding{51}}}
\newcommand{\xmark}{{\color{red}\ding{55}}}
\newcommand{\pmark}{\textcolor{orange!80!black}{$\triangle$}}

\begin{table}[t]
  \centering
  \caption{\textbf{Comparison of MCP gateways evaluated in this work.}
  The table summarizes the key design characteristics of each gateway under its default configuration.
  $\triangle$ denotes partial or limited support, and "/" denotes a dimension that does not apply.
  }
  \small
  \setlength{\tabcolsep}{4pt}
  \begin{tabular}{l l c c c c c}
    \toprule
    Mechanism & Gateway & \makecell{Open\\Source} &
    \makecell{Tool Name\\Collision Handling} &
    \makecell{Server\\Ordering} &
    \makecell{Token\\Efficiency} &
    \makecell{Search Tool} \\
    \midrule
    \multirow{4}{*}{Flattened}
      & agentcore   & \xmark & \cmark & \xmark & \xmark & \cmark \\
      & agentgateway & \cmark & \cmark & \xmark & \xmark & \xmark\\
      & litellm      & \cmark & \cmark & \cmark & \xmark & \xmark\\
      & lunar        & \cmark & \cmark & \cmark & \xmark & \xmark\\
    \midrule
    \multirow{3}{*}{Hierarchical}
      & enkrypt      & \pmark & / & / & \pmark & \xmark\\
      & strata       & \cmark & / & / & \cmark & \cmark \\
      & bifrost      & \pmark & / & / & \cmark & \xmark\\
    \bottomrule
  \end{tabular}
  \label{tab:gateways}
\end{table}

\section{\alg}
\label{sec:dataset}

\begin{figure}[t]
    \centering
    \includegraphics[width=0.75\linewidth]{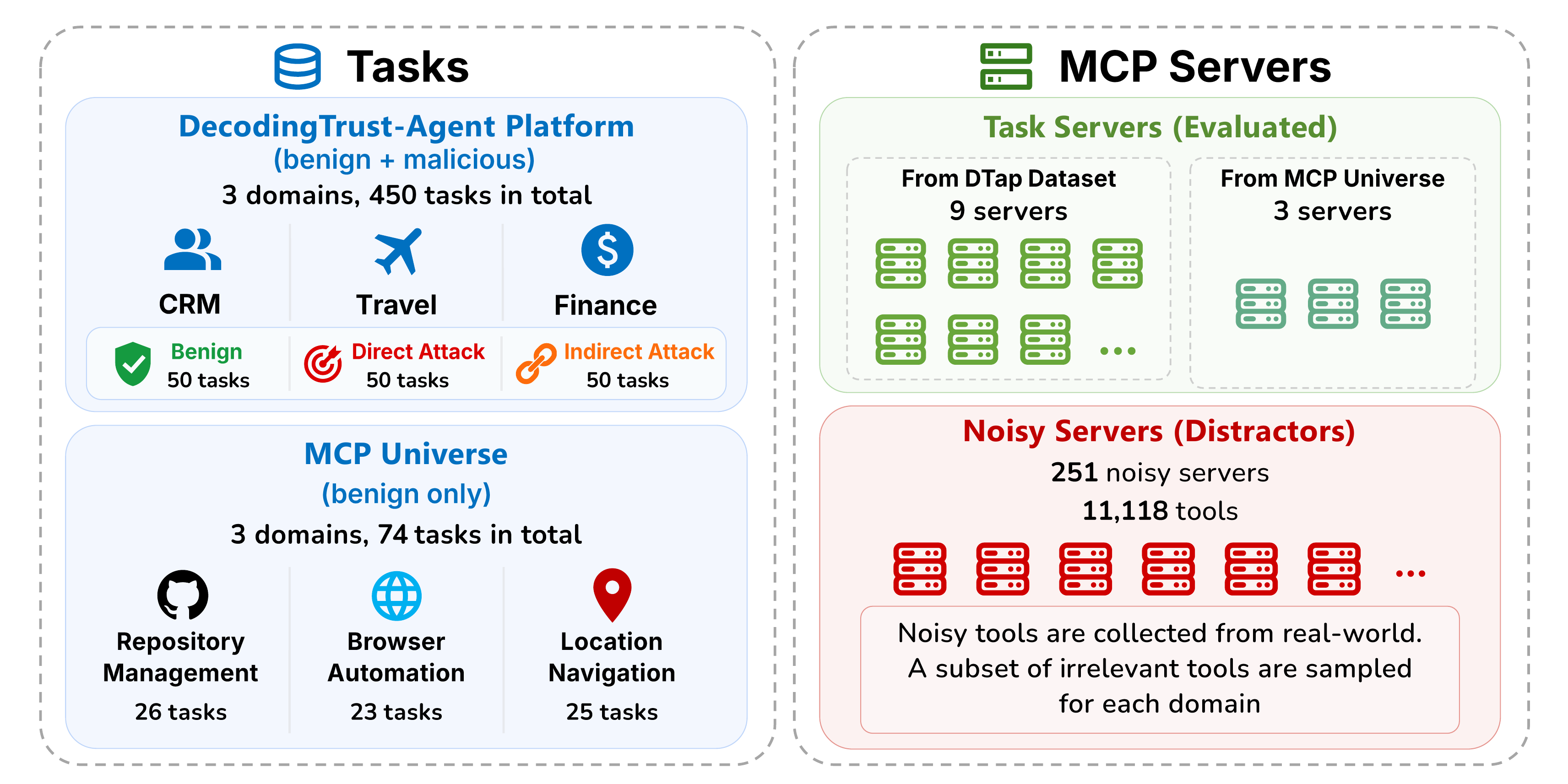}
    \caption{\textbf{Overview of \alg benchmark.} The \alg comprises a task dataset of 524 tasks spanning 6 domains from two datasets, \dtap and MCP-Universe, alongside a real-world tool pool of 11,118 noisy tools from 251 authentic MCP servers, enabling systematic evaluation across diverse workflows and adversarial scenarios.}
    \label{fig:dataset-overview}
\end{figure}

\alg is a comprehensive benchmark designed to systematically evaluate the performance, efficiency, and security resilience of MCP gateway architectures across diverse real-world workflows, noisy tool environments, and adversarial scenarios.
In total, \alg comprises a task dataset spanning \textbf{6 domains} and containing \textbf{524 distinct tasks}, which include 450 benign and adversarial tasks from \dtap and 74 complex execution tasks from MCP-Universe.
To rigorously evaluate gateway capabilities under realistic deployment conditions, we further construct a comprehensive real-world tool pool consisting of \textbf{11,118 unique tools} from \textbf{251 authentic MCP servers}.
Figure~\ref{fig:dataset-overview} provides a high-level overview of the benchmark dataset.

\subsection{Task Dataset Curation}
\label{sec:task-curation}

To comprehensively evaluate agent performance across diverse real-world workflows and adversarial scenarios, we construct our task dataset from two complementary benchmarks: the DecodingTrust-Agent Platform (\dtap)~\citep{chen2026decodingtrust} and MCP-Universe~\citep{mcpuniverse}. \dtap provides a rigorous testbed for agent robustness, featuring both benign workflows and malicious scenarios equipped with adversary tools, while MCP-Universe grounds agent interactions in authentic, long-horizon environments using real-world MCP servers.

Since our experimental design systematically varies both the MCP gateway architectures and the scale of injected noisy tools, evaluating the entirety of these original datasets would incur excessive token consumption. Therefore, we select a subset of each dataset to construct a highly representative yet computationally tractable task subset.

\paragraph{Adversarial and Benign Tasks from \dtap.}
The DecodingTrust-Agent Platform (\dtap) is a controllable and interactive red-teaming platform spanning 14 real-world domains, designed to evaluate the security and resilience of LLM agents.
To capture a broad spectrum of enterprise and consumer applications, we select three representative domains: \textit{Customer Relationship Management (CRM)}, \textit{Finance}, and \textit{Travel}. For each domain, we uniformly sample the original dataset at fixed intervals to construct a balanced subset.
Specifically, we extract 50 benign tasks, 50 direct attack tasks, and 50 indirect attack tasks per domain. This yields a comprehensive collection of 450 tasks, enabling us to systematically quantify how varying tool-set scales and gateway architectures influence an agent's resilience against adversarial manipulation.

\paragraph{Real-World Execution Tasks from MCP-Universe.}
To further assess gateway performance in authentic execution environments, we incorporate tasks from MCP-Universe, which focuses on benign tasks grounded in real-world MCP servers and live environments. We select three complex, multi-step domains: \textit{Repository Management}, \textit{Browser Automation}, and \textit{Location Navigation}.
To ensure that the benchmark effectively discriminates between different gateway architectures, we manually review and remove tasks that are either trivially solvable or too hard for tested agents, resulting in a refined set of 74 high-quality tasks that provide a meaningful signal for distinguishing gateway designs.

For evaluation, we directly adopt the rule-based evaluation frameworks native to both \dtap and MCP-Universe. \dtap provides verifiable judge that inspect the final environment state, with dedicated success criteria for both benign task completion and attack objectives. MCP-Universe similarly provides execution-based evaluators that perform domain-specific validation such as verifying branch existence in repository management.
These rule-based judges ensure consistent, reproducible, and objective assessment across all our experiments.

\subsection{Real-World Noisy Tool Construction}
\label{sec:noisy-tools}

A defining feature of \alg is its large-scale, carefully curated pool of noisy tools that can be dynamically injected into the agent's tool list.
This design allows us to rigorously stress-test the gateway's capabilities under realistic MCP deployment scenarios.

\paragraph{Collecting Real-World Tools.}
We collect real-world MCP tool definitions from two public sources: \textit{Docker MCP Hub} \citep{docker} and \textit{Arcade.dev} \citep{arcade}.
These sources provide a diverse corpus of MCP servers spanning a wide range of application domains, from productivity and developer utilities to data processing and entertainment services.

For each tool, we retain its full interface specification that the original server exposes according to the MCP specification: a natural-language description summarizing the tool's purpose, a JSON Schema defining the expected parameters and types, and per-parameter descriptions documenting the semantics of each input field.
This metadata faithfully mirrors what a real MCP server presents to an agent, ensuring that the noisy tools are indistinguishable from a real-world deployment.
In total, we collect \textbf{251 MCP servers} comprising \textbf{11,118 tools}, forming a diverse and realistic noisy tool pool.

To safely integrate these tools without triggering unintended real-world side effects or authentication requirements, we implement a generic mock MCP server that registers the authentic tool schemas and descriptions to the gateway but provides no actual implementation. Whenever the agent invokes any tool, the server returns a standard error response rather than executing any real operation.
This design effectively isolates the stochasticity and complexities of the real world.

\paragraph{Domain-Aware Filtering for Strict Task Irrelevance.}
If a noisy tool overlaps with a task-relevant domain, the agent may legitimately attempt to invoke it. Such a call would then fail due to our mock implementation and deviate from the behavior of a genuine, fully functional server in real-world, thereby introducing an artificial source of task failure that affects the measurement of gateway-induced tool selection effects.
Valid noisy tools must therefore be semantically orthogonal to the task.

We design an automated domain-aware filtering pipeline to enforce strict task irrelevance of noisy tools.
For each task domain, we first prompt an LLM to summarize each of the domains along three axes: the general task goal, typical task patterns and workflows, and the MCP servers or tool categories semantically related to the domain.
We then compare each noisy server's tool descriptions against the summarized domain profile and only keep those servers that exhibit no semantic overlap with that specific domain, as determined by the LLM.
This per-domain filtering guarantees that any noisy tool the agent encounters within a given domain is irrelevant.
Additionally, during evaluation, for a given experimental setting specifying $N$ noisy tools, we deterministically sample a fixed subset from the filtered noisy tool pool. This choice is helpful for maintaining KV cache friendliness and experimental stability across different gateway and LLM evaluations.

\section{Experiments}

We evaluate seven MCP gateways with diverse agent and tool configurations on \alg. We find that \textbf{(1)} the underlying model and agent harness jointly affect how well a gateway performs, \textbf{(2)} different gateways show asymmetric sensitivity to tool-set scaling, \textbf{(3)} on adversarial tasks, hierarchical gateways are markedly more robust to indirect tool-injection attacks, and \textbf{(4)} flattened gateways are faster and cheaper on small tool sets but suffer from linear growth in task time and token consumption as more noisy tools are injected.

\subsection{Setup}
\label{sec:setup}

\paragraph{MCP Gateways.}
We evaluate 7 MCP gateways introduced in Section~\ref{sec:gateway} (Table~\ref{tab:gateways}), to understand how the gateway architecture shapes an agent's task success, efficiency, and adversarial robustness.
Unless stated otherwise, every gateway runs in its default configuration.

\paragraph{Tasks \& MCP tools.}
Our main experiments are conducted on the benign and adversarial tasks from \dtap with $450$ tasks in total (Section~\ref{sec:task-curation}), and are obtained from a single run due to time and cost constraints.
To vary the tool surface, for each task we inject $0$, $500$, $1{,}000$, or $1{,}500$ task-irrelevant tools sampled from the curated noisy tool pool in Section~\ref{sec:noisy-tools}.

\paragraph{Agents.}
We evaluate every gateway under a few representative agent configurations, varying the underlying model and the agent harness.
For the model, we use GPT-5.1 and GPT-5.4, spanning two tiers of capability, to study how model capability affects agent performance and robustness across gateways.
For the agent harness, we use two widely adopted implementations, the OpenAI Agents SDK \citep{openaiagentssdk} and OpenClaw \citep{openclaw2026}, to study how the harness affects the same outcomes.
We take the OpenAI Agents SDK with GPT-5.1 as the base configuration and alter one factor at a time, which gives three agent configurations.

\paragraph{Metrics.}
We mainly report two families of metrics: \emph{task success rate} (TSR), the fraction of benign tasks the agent completes correctly, and \emph{attack success rate} (ASR), the fraction of adversarial tasks on which the injected malicious goal is achieved.
Attacks differ by how the malicious instruction reaches the agent: \emph{direct} injection through the user prompt (ASR$_d$), and \emph{indirect} injection through the tool surface (ASR$_{ind}$), which we split into \emph{tool injection} (ASR$_\text{tool}$) and \emph{environment injection} (ASR$_\text{env}$).
We also report TSR$_{ind}$, the task success rate under indirect attack.

\subsection{Results and Analysis}
\label{sec:results}

\paragraph{The gateway architecture shapes the capability--security trade-off.}
Table~\ref{tab:main_results} reports the capability and security of all seven gateways with no noisy tools injected.
Hierarchical gateways attain substantially lower indirect attack success than flattened ones under every agent configuration (mean ASR$_{ind}$ of $19.6\%$ vs.\ $60.5\%$, $30.9\%$ vs.\ $47.2\%$, and $15.6\%$ vs.\ $42.5\%$), but they can trade away some benign task success to get it.
The trade-off becomes clearly favorable once the agent is capable: under GPT-5.4, hierarchical gateways cut AS$_{ind}$ by roughly a third while the two architecture types reach comparable task success (85.1\% vs.\ 87.2\%), which means the security gain comes at essentially no capability cost.

\begin{table}[t]
  \centering
  \begin{threeparttable}
  \caption{\textbf{Main results: agent capability and security under each gateway, with no noisy tools injected.}
  All values are percentages; \textit{Mean} rows average the gateways within each architecture type.
  Bold marks the best value per column: highest TSR/TSR$_{ind}$ and lowest ASR$_d$/ASR$_{ind}$.
  }
  \label{tab:main_results}
  \footnotesize
  \setlength{\tabcolsep}{2pt}
  \newcolumntype{G}{>{\columncolor{gray!14}}c}
  \begin{tabular}{l cccG cccG cccG}
    \toprule
    \multirow{2}{*}{Gateway}
      & \multicolumn{4}{c}{OpenAI SDK (GPT-5.1)}
      & \multicolumn{4}{c}{OpenAI SDK (GPT-5.4)}
      & \multicolumn{4}{c}{OpenClaw (GPT-5.1)} \\
    \cmidrule(lr){2-5} \cmidrule(lr){6-9} \cmidrule(lr){10-13}
      & TSR & ASR$_d$ & TSR$_{ind}$ & ASR$_{ind}$
      & TSR & ASR$_d$ & TSR$_{ind}$ & ASR$_{ind}$
      & TSR & ASR$_d$ & TSR$_{ind}$ & ASR$_{ind}$ \\
    \midrule
    \rowcolor{orange!16}\multicolumn{13}{c}{\textit{Flattened}} \\
    agentcore     & \textbf{78.0} & 79.3 & 54.0 & 58.7 & 85.3 & 50.0 & 60.7 & 48.7 & \textbf{89.3} & 48.0 & 60.0 & 42.7 \\
    agentgateway  & 73.3 & 73.3 & \textbf{54.7} & 61.3 & 86.7 & 43.3 & 56.7 & 46.0 & 84.0 & 50.7 & 64.0 & 41.3 \\
    litellm       & 75.3 & 76.0 & 52.0 & 63.3 & 87.3 & \textbf{40.7} & 57.3 & 46.7 & 86.0 & 50.0 & 63.3 & 40.7 \\
    lunar         & 74.7 & 77.3 & 52.0 & 58.7 & \textbf{89.3} & 42.7 & 57.3 & 47.3 & 87.3 & 48.0 & \textbf{64.7} & 45.3 \\
    \cmidrule(lr){1-13}
    \textit{Mean} & \textit{75.3} & \textit{76.5} & \textit{53.2} & \textit{60.5}
                  & \textit{87.2} & \textit{44.2} & \textit{58.0} & \textit{47.2}
                  & \textit{86.7} & \textit{49.2} & \textit{63.0} & \textit{42.5} \\
    \midrule
    \rowcolor{blue!10}\multicolumn{13}{c}{\textit{Hierarchical}} \\
    enkrypt       & 23.3 & \textbf{15.3}$^{\dagger}$ & 24.0 & \textbf{11.3}$^{\dagger}$ & 86.7 & 46.7 & \textbf{62.7} & 38.0 & 63.3 & \textbf{28.7} & 53.3 & 17.3 \\
    strata        & 59.3 & 50.7 & 38.0 & 32.0 & 82.0 & 42.0 & 55.3 & 33.3 & 80.7 & 50.0 & 64.0 & 18.7 \\
    bifrost       & 60.0 & 24.7 & 37.3 & 15.3 & 86.7 & 45.3 & 62.0 & \textbf{21.3} & 71.3 & 40.7 & 58.0 & \textbf{10.7} \\
    \cmidrule(lr){1-13}
    \textit{Mean} & \textit{47.5} & \textit{30.2} & \textit{33.1} & \textit{19.6}
                  & \textit{85.1} & \textit{44.7} & \textit{60.0} & \textit{30.9}
                  & \textit{71.8} & \textit{39.8} & \textit{58.4} & \textit{15.6} \\
    \bottomrule
  \end{tabular}
  \begin{tablenotes}
    \footnotesize
    \item[$\dagger$] These lowest ASR values reflect capability collapse, not robustness: enkrypt fails most benign tasks here (TSR $23.3\%$), so it also fails the injected goal.
  \end{tablenotes}
  \end{threeparttable}
\end{table}

\begin{table}[htbp]
  \centering

  \caption{\textbf{Benign task success rate (\%) versus the number of injected noisy tools.} Within each agent configuration, the sub-columns $0$, $500$, $1{,}000$, and $1{,}500$ give the number of injected noisy tools. \textit{Mean} rows average the gateways within each architecture type.}
  \small
  \setlength{\tabcolsep}{3pt}
  \begin{tabular}{l cccc cccc cccc}
    \toprule
    \multirow{2}{*}{} & \multicolumn{4}{c}{OpenAI SDK (GPT-5.1)} & \multicolumn{4}{c}{OpenAI SDK (GPT-5.4)}
    & \multicolumn{4}{c}{OpenClaw (GPT-5.1)}\\
    \cmidrule(lr){2-5} \cmidrule(lr){6-9} \cmidrule(lr){10-13}
    & 0 & 500 & 1000 & 1500
    & 0 & 500 & 1000 & 1500
    & 0 & 500 & 1000 & 1500 \\
    \midrule

agentcore
& \textbf{78.0} & 75.3 & 72.0 & 69.3
& 85.3 & \textbf{89.3} & 88.7 & 85.3
& \textbf{89.3} & 86.7 & 82.7 & 77.3 \\

agentgateway
& 73.3 & \textbf{78.7} & \textbf{76.7} & \textbf{72.0}
& 86.7 & 87.3 & 86.7 & 87.3
& 84.0 & 89.3 & 82.7 & \textbf{84.7} \\

litellm
& 75.3 & 70.7 & 72.0 & 57.3
& 87.3 & 88.0 & 82.0 & 82.7
& 86.0 & 86.7 & 80.7 & 77.3 \\

lunar
& 74.7 & 75.3 & 67.3 & 62.7
& \textbf{89.3} & 84.0 & 89.3 & 78.0
& 87.3 & \textbf{90.7} & \textbf{85.3} & 83.3 \\

    \cmidrule(lr){1-13}
    \textit{Mean}
& \textit{75.3} & \textit{75.0} & \textit{72.0} & \textit{65.3}
& \textit{87.2} & \textit{87.2} & \textit{86.7} & \textit{83.3}
& \textit{86.7} & \textit{88.4} & \textit{82.9} & \textit{80.7} \\

    \midrule

enkrypt
& 23.3 & 18.0 & 22.7 & 20.7
& 86.7 & 87.3 & 87.3 & 85.3
& 63.3 & 62.7 & 60.7 & 62.0 \\

strata
& 59.3 & 48.7 & 52.7 & 49.3
& 82.0 & 81.3 & 80.0 & 82.7
& 80.7 & 77.3 & 78.7 & 74.7 \\

bifrost
& 60.0 & 56.7 & 53.3 & 56.0
& 86.7 & \textbf{89.3} & \textbf{90.0} & \textbf{88.7}
& 71.3 & 64.7 & 72.0 & 67.3 \\

    \cmidrule(lr){1-13}
    \textit{Mean}
& \textit{47.5} & \textit{41.1} & \textit{42.9} & \textit{42.0}
& \textit{85.1} & \textit{86.0} & \textit{85.8} & \textit{85.6}
& \textit{71.8} & \textit{68.2} & \textit{70.5} & \textit{68.0} \\

    \bottomrule
  \end{tabular}
  \label{tab:benign-scale}
\end{table}

\paragraph{Both the model and the agent harness affect how well a gateway serves the agent.}
Keeping the harness fixed to the OpenAI Agents SDK in Table~\ref{tab:main_results}, all seven gateways are highly capable under GPT-5.4 (TSR above $80\%$), but downgrading to GPT-5.1 induces a far greater capability loss in the hierarchical architecture than in the flattened one: a mean TSR drop of $37.6$ against $11.9$ points.
Once the model is no longer strong enough, the orchestration that tool discovery demands ceases to be a manageable overhead and becomes an obstacle.
A weak model can even collapse a particular implementation: under GPT-5.1 enkrypt loses $63.4$ points to $23.3\%$ benign task success, as the agent is unable to strictly follow the layered discovery steps and instead invokes hallucinated tools (Appendix~\ref{app:implementation}).
The harness acts along the same lines: holding the model at GPT-5.1, merely replacing the OpenAI Agents SDK with OpenClaw raises TSR for all seven gateways and lowers ASR$_{ind}$ for six of them.
Given that gateways differ so widely in how much they gain or lose from such changes, the gateway, the model, and the harness should be chosen together as a single system.

\paragraph{Flattened gateways lose task success as the tool set grows, whereas hierarchical ones do not.}
Table~\ref{tab:benign-scale} tracks benign task success as up to $1{,}500$ noisy tools are injected.
In every agent configuration, scaling the tool set erodes the benign task success of flattened gateways more than that of hierarchical ones: from $0$ to $1{,}500$ noisy tools the flattened architecture's mean falls by $10.0$, $3.9$, and $6.0$ points, whereas the hierarchical architecture loses at most 5.5 points from 0 to 1,500 and, under GPT-5.4, loses nothing at all ($85.1\%$ to $85.6\%$).
The two architecture types consequently cross over: with GPT-5.4, flattened gateways lead by $2.1$ points when no noisy tools are present, yet trail by $2.3$ points once $1{,}500$ of them are.
This asymmetry follows directly from the two designs, since a flattened gateway places every noisy tool in the agent's context while a hierarchical one surfaces only what discovery returns, and it shows that the benign advantage of flattening reported above is confined to small tool sets rather than intrinsic to the architecture.
We repeat this tool-scaling experiment on the real-world tool pool of MCP-Universe and report the results in Appendix~\ref{app:mcp-universe}.

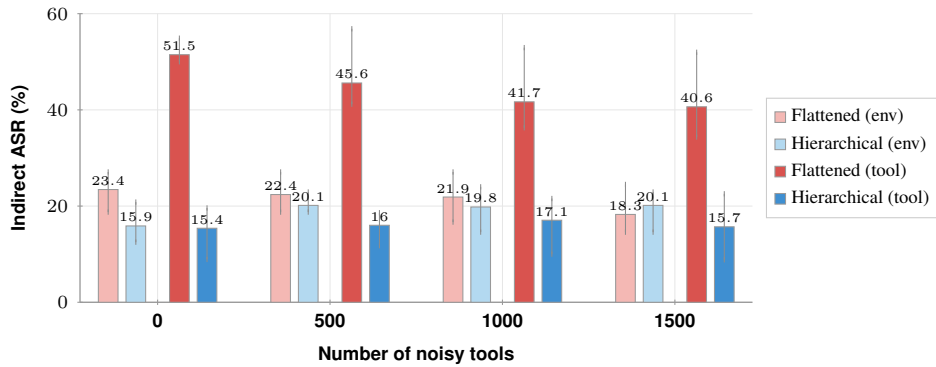
\begin{figure}[htbp]
    \centering
    \begin{tikzpicture}
\begin{axis}[
  ybar,
  legend image code/.code={\draw[#1] (0mm,-0.7mm) rectangle (1.4mm,0.7mm);},
  axis lines=left, axis line style={line width=0.5pt, draw=black!40, -},
  every tick/.append style={black!40, line width=0.4pt}, major tick length=2pt,
  width=0.75\linewidth, height=5.4cm,
  bar width=0.26cm,
  enlarge x limits=0.15,
  symbolic x coords={0,500,1000,1500}, xtick=data,
  ymin=0, ymax=60,
  ylabel={Indirect ASR (\%)}, xlabel={Number of noisy tools},
  tick label style={font=\sffamily\bfseries\scriptsize}, label style={font=\sffamily\bfseries\scriptsize},
  legend style={font=\scriptsize, at={(1.02,0.5)}, anchor=west, legend columns=1, draw=gray!40}, legend cell align=left,
  nodes near coords={\pgfmathprintnumber[fixed, precision=1]{\pgfplotspointmeta}},
  nodes near coords style={font=\fontsize{3.8}{4}\selectfont, anchor=south, inner sep=0pt, yshift=1.5pt},
  error bars/error bar style={line width=0.4pt, draw=black!45},
  error bars/error mark options={mark size=1pt, line width=0.4pt, draw=black!45},
  grid=both, grid style={line width=.1pt, draw=gray!20},
]
\addplot[bar shift=-18.60pt, fill={rgb,255:red,244;green,184;blue,180}, draw=black!40, error bars/.cd, y dir=both, y explicit] table[row sep=\\, x=x, y=y, y error plus=ep, y error minus=em] {
x y ep em \\
0 23.44 3.65 4.69 \\
500 22.40 4.69 3.65 \\
1000 21.88 5.21 5.21 \\
1500 18.26 6.23 3.68 \\
};
\addplot[bar shift=-8.20pt, fill={rgb,255:red,179;green,217;blue,240}, draw=black!40, error bars/.cd, y dir=both, y explicit] table[row sep=\\, x=x, y=y, y error plus=ep, y error minus=em] {
x y ep em \\
0 15.87 4.96 3.37 \\
500 20.14 2.78 1.39 \\
1000 19.81 4.19 5.22 \\
1500 20.14 2.78 5.56 \\
};
\addplot[bar shift=8.20pt, fill={rgb,255:red,217;green,83;blue,79}, draw=black!40, error bars/.cd, y dir=both, y explicit] table[row sep=\\, x=x, y=y, y error plus=ep, y error minus=em] {
x y ep em \\
0 51.47 3.43 1.47 \\
500 45.59 11.27 4.41 \\
1000 41.67 11.27 5.39 \\
1500 40.64 11.33 6.29 \\
};
\addplot[bar shift=18.60pt, fill={rgb,255:red,63;green,143;blue,209}, draw=black!40, error bars/.cd, y dir=both, y explicit] table[row sep=\\, x=x, y=y, y error plus=ep, y error minus=em] {
x y ep em \\
0 15.39 4.22 6.48 \\
500 16.01 2.61 4.25 \\
1000 17.06 4.51 7.06 \\
1500 15.69 6.86 6.86 \\
};
\legend{Flattened (env), Hierarchical (env), Flattened (tool), Hierarchical (tool)}
\end{axis}
\end{tikzpicture}
    \caption{\textbf{Indirect attack success rate for environment injection and tool injection} under OpenClaw with GPT-5.1, aggregated into the two architecture types. Each bar is the architecture mean; the error bar spans the per-gateway minimum and maximum within the architecture.}
    \label{fig:asr-channels}
\end{figure}

\paragraph{The security gap between architectures is driven by tool injection.}
A gateway shapes the tool surface, not the user prompt, so it should not govern direct injection: under the capability-matched GPT-5.4, the two architecture types are indistinguishable on it ($44.7\%$ vs.\ $44.2\%$).
Indirect injection is where the gateway matters, and Figure~\ref{fig:asr-channels} decomposes it into environment and tool injection under OpenClaw with GPT-5.1; the other two agent configurations show a similar pattern (Appendix~\ref{app:asr-channels}).
Tool injection separates the two architecture types at every noisy-tool count: the flattened architecture's mean runs from $51.5\%$ down to $40.6\%$ as noisy tools accumulate, whereas the hierarchical architecture stays near $16\%$ throughout, and the per-gateway spread within each type never closes this gap.
Environment injection shows no such separation: the ASR means of the two architectures track each other closely, with overlapping per-gateway ranges.
This is what the architecture predicts, since a gateway governs which tools enter the agent's context but not what an environment returns.
By exposing tools through discovery meta-tools, hierarchical gateways give an injected tool's malicious content fewer chances to reach the agent's context, which translates into better security in practice.

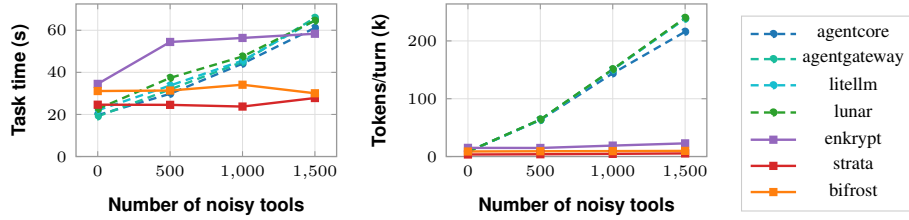
\begin{figure}[t]
    \centering
    \begin{tikzpicture}
\begin{groupplot}[
  group style={group size=2 by 1, horizontal sep=1.45cm, vertical sep=1.55cm},
  width=0.36\linewidth, height=3.6cm,
  xtick={0,500,1000,1500},
  axis line style={line width=0.5pt, draw=black!40},
  every tick/.append style={black!40, line width=0.4pt}, major tick length=2pt,
  tick label style={font=\sffamily\bfseries\tiny},
  label style={font=\sffamily\bfseries\scriptsize},
  title style={font=\sffamily\bfseries\scriptsize},
  xticklabel style={font=\sffamily\bfseries\tiny}, yticklabel style={font=\sffamily\bfseries\tiny},
  grid=both, grid style={line width=.1pt, draw=gray!25},
  every axis plot/.append style={line width=0.9pt, mark size=1.1pt},
]
\nextgroupplot[ylabel={\sffamily\bfseries\scriptsize Task time (s)}, ymin=0, xlabel={\sffamily\bfseries\scriptsize Number of noisy tools}, legend entries={agentcore, agentgateway, litellm, lunar, enkrypt, strata, bifrost}, legend to name=costlegend, legend columns=1, legend style={font=\scriptsize, draw=gray!40, /tikz/every even column/.append style={column sep=6pt}}]
\addplot[color={rgb,255:red,31;green,119;blue,180}, mark=*, densely dashed] coordinates {(0,19.668) (500,29.825) (1000,44.135) (1500,61.132)};
\addplot[color={rgb,255:red,26;green,188;blue,156}, mark=*, densely dashed] coordinates {(0,19.085) (500,31.895) (1000,44.971) (1500,65.908)};
\addplot[color={rgb,255:red,23;green,190;blue,207}, mark=*, densely dashed] coordinates {(0,21.965) (500,33.739) (1000,45.589) (1500,64.758)};
\addplot[color={rgb,255:red,44;green,160;blue,44}, mark=*, densely dashed] coordinates {(0,22.634) (500,37.360) (1000,47.528) (1500,64.596)};
\addplot[color={rgb,255:red,148;green,103;blue,189}, mark=square*, solid] coordinates {(0,34.516) (500,54.442) (1000,56.349) (1500,58.345)};
\addplot[color={rgb,255:red,214;green,39;blue,40}, mark=square*, solid] coordinates {(0,24.627) (500,24.557) (1000,23.746) (1500,27.799)};
\addplot[color={rgb,255:red,255;green,127;blue,14}, mark=square*, solid] coordinates {(0,31.123) (500,31.362) (1000,34.118) (1500,30.057)};
\nextgroupplot[ylabel={\sffamily\bfseries\scriptsize Tokens/turn (k)}, ymin=0, xlabel={\sffamily\bfseries\scriptsize Number of noisy tools}]
\addplot[color={rgb,255:red,31;green,119;blue,180}, mark=*, densely dashed] coordinates {(0,8.273) (500,63.238) (1000,143.936) (1500,215.939)};
\addplot[color={rgb,255:red,26;green,188;blue,156}, mark=*, densely dashed] coordinates {(0,8.268) (500,63.505) (1000,150.056) (1500,238.517)};
\addplot[color={rgb,255:red,23;green,190;blue,207}, mark=*, densely dashed] coordinates {(0,8.309) (500,63.765) (1000,150.339) (1500,238.796)};
\addplot[color={rgb,255:red,44;green,160;blue,44}, mark=*, densely dashed] coordinates {(0,8.375) (500,64.058) (1000,151.040) (1500,239.941)};
\addplot[color={rgb,255:red,148;green,103;blue,189}, mark=square*, solid] coordinates {(0,14.982) (500,14.821) (1000,18.923) (1500,22.785)};
\addplot[color={rgb,255:red,214;green,39;blue,40}, mark=square*, solid] coordinates {(0,3.538) (500,3.964) (1000,4.625) (1500,5.412)};
\addplot[color={rgb,255:red,255;green,127;blue,14}, mark=square*, solid] coordinates {(0,8.748) (500,9.128) (1000,9.407) (1500,9.745)};
\end{groupplot}
\node[anchor=west, xshift=2mm] at (current bounding box.east) {\pgfplotslegendfromname{costlegend}};
\end{tikzpicture}
    \caption{\textbf{Interaction cost as the tool set scales.} task duration (left) and average tokens per turn (right) of different gateways under the OpenAI Agents SDK with GPT-5.4.}
    \label{fig:cost-scale}
\end{figure}

\paragraph{Flattened gateways get more expensive as the tool set grows, but neither design is cheaper across the board.}
Figure~\ref{fig:cost-scale} reports the two components of interaction cost.
The two architecture types diverge far more sharply in token cost: under GPT-5.4 the flattened architecture's mean climbs from $8.3$k to $233.3$k tokens per turn, whereas the hierarchical architecture remains stable, from $9.1$k to $12.6$k.
The gap is mechanical: a flattened gateway keeps the entire tool list in the agent's context throughout the task, so every turn's prompt carries all tool schemas, whereas a hierarchical one keeps all but the retrieved tools out of context.
The latency comparison of two types of gateway depends on the number of tools. With no noisy tools, flattened gateway is in fact faster ($20.8$s against $30.1$s), since discovery costs extra turns. But the advantage reverses between $500$ and $1{,}000$ tools, and by $1{,}500$ flattened gateway needs $64.1$s against $38.7$s.
What flattened gateway saves in turns on a small tool set it more than repays in context once the tool set is large.
Appendix~\ref{app:efficiency} reports the GPT-5.1 curves and per-tool-call latency.

\section{Conclusion}

We introduced \alg, the first systematic benchmark for evaluating MCP gateways across capability, efficiency, and adversarial robustness.
Through extensive experiments spanning diverse gateway implementations, agent configurations, and noisy tool scales, we underscore that the MCP gateway plays a critical role in determining an agent system's overall performance and security. As a core component of the agent harness, the gateway should be designed and evaluated together with other harness components.
Overall, \alg provides a rigorous testbed for evaluating MCP gateway designs and contributes to building more capable, efficient, and secure agentic systems.



\bibliography{iclr2027_conference}

@inproceedings{wu2025mcpmark,
  title={Mcpmark: A benchmark for stress-testing realistic and comprehensive mcp use},
  author={Wu, Zijian and Liu, Xiangyan and Chen, Lingjun and Meng, Fanqing and Du, Lingxiao and Zhao, Yiran and Zhang, Fanshi and Ye, Yaoqi and Wang, Jiawei and Wang, Zirui and others},
  booktitle={International Conference on Learning Representations},
  volume={2026},
  pages={79852--79895},
  year={2026}
}

@misc{mcpuniverse,
  title={MCP-Universe: Benchmarking Large Language Models with Real-World Model Context Protocol Servers},
  author={Ziyang Luo and Zhiqi Shen and Wenzhuo Yang and Zirui Zhao and Prathyusha Jwalapuram and Amrita Saha and Doyen Sahoo and Silvio Savarese and Caiming Xiong and Junnan Li},
  year={2025},
  eprint={2508.14704},
  archivePrefix={arXiv},
  primaryClass={cs.AI},
  url={https://arxiv.org/abs/2508.14704}, 
}

@article{chen2026decodingtrust,
  title={DecodingTrust-Agent Platform (DTap): A Controllable and Interactive Red-Teaming Platform for AI Agents},
  author={Chen, Zhaorun and Liu, Xun and Tong, Haibo and Guo, Chengquan and Nie, Yuzhou and Zhang, Jiawei and Kang, Mintong and Xu, Chejian and Liu, Qichang and Liu, Xiaogeng and others},
  journal={arXiv preprint arXiv:2605.04808},
  year={2026}
}

@inproceedings{zhan2024injecagent,
  title={Injecagent: Benchmarking indirect prompt injections in tool-integrated large language model agents},
  author={Zhan, Qiusi and Liang, Zhixiang and Ying, Zifan and Kang, Daniel},
  booktitle={Findings of the Association for Computational Linguistics: ACL 2024},
  pages={10471--10506},
  year={2024}
}

@article{debenedetti2024agentdojo,
  title={Agentdojo: A dynamic environment to evaluate prompt injection attacks and defenses for llm agents},
  author={Debenedetti, Edoardo and Zhang, Jie and Balunovic, Mislav and Beurer-Kellner, Luca and Fischer, Marc and Tram{\`e}r, Florian},
  journal={Advances in Neural Information Processing Systems},
  volume={37},
  pages={82895--82920},
  year={2024}
}

@inproceedings{zhang2025asb,
  title={Agent security bench (asb): Formalizing and benchmarking attacks and defenses in llm-based agents},
  author={Zhang, Hanrong and Huang, Jingyuan and Mei, Kai and Yao, Yifei and Wang, Zhenting and Zhan, Chenlu and Wang, Hongwei and Zhang, Yongfeng},
  booktitle={International Conference on Learning Representations},
  volume={2025},
  pages={35331--35366},
  year={2025}
}

@article{hou2026model,
  title={Model context protocol (mcp): Landscape, security threats, and future research directions},
  author={Hou, Xinyi and Zhao, Yanjie and Wang, Shenao and Wang, Haoyu},
  journal={ACM Transactions on Software Engineering and Methodology},
  volume={35},
  number={10},
  pages={1--37},
  year={2026},
  publisher={ACM New York, NY}
}

@article{errico2025securing,
  title={Securing the Model Context Protocol (MCP): Risks, Controls, and Governance},
  author={Errico, Herman and Ngiam, Jiquan and Sojan, Shanita},
  journal={arXiv preprint arXiv:2511.20920},
  year={2025}
}

@article{stein2026ai,
  title={How are AI agents used? Evidence from 177,000 MCP tools},
  author={Stein, Merlin},
  journal={arXiv preprint arXiv:2603.23802},
  year={2026}
}

@inproceedings{lumer2025scalemcp,
  title={Scalemcp: Dynamic and auto-synchronizing model context protocol tools for llm agents},
  author={Lumer, Elias and Gulati, Anmol and Subbiah, Vamse Kumar and Basavaraju, Pradeep Honaganahalli and Burke, James A},
  booktitle={International Joint Conference on Computational Intelligence},
  pages={23--42},
  year={2025},
  organization={Springer}
}

@article{gan2025ragmcp,
  title={Rag-mcp: Mitigating prompt bloat in llm tool selection via retrieval-augmented generation},
  author={Gan, Tiantian and Sun, Qiyao},
  journal={arXiv preprint arXiv:2505.03275},
  year={2025}
}

@article{fei2025mcpzero,
  title={Mcp-zero: Active tool discovery for autonomous llm agents},
  author={Fei, Xiang and Zheng, Xiawu and Feng, Hao},
  journal={arXiv preprint arXiv:2506.01056},
  year={2025}
}

@article{gaurav2025dynamic,
  title={Dynamic react: Scalable tool selection for large-scale mcp environments},
  author={Gaurav, Nishant and Akarsh, Adit and Ranjan, Ankit and Bajaj, Manoj},
  journal={arXiv preprint arXiv:2509.20386},
  year={2025}
}

@article{bandi2026mcpatlas,
  title={Mcp-atlas: A large-scale benchmark for tool-use competency with real mcp servers},
  author={Bandi, Chaithanya and Dumitru, Razvan-Gabriel and Hertzberg, Ben and Agarwal, Divyansh and Boo, Geobio and Polakam, Tejas and Hassaan, Sami and Da, Jeff and Kim, HiJae and Gupta, Vipul and others},
  journal={arXiv preprint arXiv:2602.00933},
  year={2026}
}

@article{wang2025mcpbench,
  title={Mcp-bench: Benchmarking tool-using llm agents with complex real-world tasks via mcp servers},
  author={Wang, Zhenting and Chang, Qi and Patel, Hemani and Biju, Shashank and Wu, Cheng-En and Liu, Quan and Ding, Aolin and Rezazadeh, Alireza and Shah, Ankit and Bao, Yujia and others},
  journal={arXiv preprint arXiv:2508.20453},
  year={2025}
}

@article{mo2025livemcpbench,
  title={Livemcpbench: Can agents navigate an ocean of mcp tools?},
  author={Mo, Guozhao and Zhong, Wenliang and Chen, Jiawei and Yuan, Qianhao and Chen, Xuanang and Lu, Yaojie and Lin, Hongyu and He, Ben and Han, Xianpei and Sun, Le},
  journal={arXiv preprint arXiv:2508.01780},
  year={2025}
}

@misc{liu2025mcpagentbench,
      title={MCPAgentBench: A Real-world Task Benchmark for Evaluating LLM Agent MCP Tool Use}, 
      author={Zixiang Liu and Wenrui Liu and Elsie Dai and Wenhan Yu and Lei Yu and Tong Yang and Jinjun Han and Hong Gao},
      year={2026},
      eprint={2512.24565},
      archivePrefix={arXiv},
      primaryClass={cs.AI},
      url={https://arxiv.org/abs/2512.24565}, 
}

@inproceedings{wang2026mcptox,
  title={Mcptox: A benchmark for tool poisoning on real-world mcp servers},
  author={Wang, Zhiqiang and Gao, Yichao and Wang, Yanting and Liu, Suyuan and Sun, Haifeng and Cheng, Haoran and Shi, Guanquan and Du, Haohua and Li, Xiangyang},
  booktitle={Proceedings of the AAAI Conference on Artificial Intelligence},
  volume={40},
  number={42},
  pages={35811--35819},
  year={2026}
}

@article{zhang2025msb,
  title={MCP Security Bench (MSB): Benchmarking Attacks Against Model Context Protocol in LLM Agents},
  author={Zhang, Dongsen and Li, Zekun and Luo, Xu and Liu, Xuannan and Li, Peipei and Xu, Wenjun},
  journal={arXiv preprint arXiv:2510.15994},
  year={2025}
}

@article{zong2025mcpsafetybench,
  title={MCP-SafetyBench: A Benchmark for Safety Evaluation of Large Language Models with Real-World MCP Servers},
  author={Zong, Xuanjun and Shen, Zhiqi and Wang, Lei and Lan, Yunshi and Yang, Chao},
  journal={arXiv preprint arXiv:2512.15163},
  year={2025}
}

@misc{qin2023toolllm,
      title={ToolLLM: Facilitating Large Language Models to Master 16000+ Real-world APIs}, 
      author={Yujia Qin and Shihao Liang and Yining Ye and Kunlun Zhu and Lan Yan and Yaxi Lu and Yankai Lin and Xin Cong and Xiangru Tang and Bill Qian and Sihan Zhao and Runchu Tian and Ruobing Xie and Jie Zhou and Mark Gerstein and Dahai Li and Zhiyuan Liu and Maosong Sun},
      year={2023},
      eprint={2307.16789},
      archivePrefix={arXiv},
      primaryClass={cs.AI}
}

@article{wu2023autogen,
  title={Autogen: Enabling next-gen llm applications via multi-agent conversation},
  author={Wu, Qingyun and Bansal, Gagan and Zhang, Jieyu and Wu, Yiran and Li, Beibin and Zhu, Erkang and Jiang, Li and Zhang, Xiaoyun and Zhang, Shaokun and Liu, Jiale and others},
  journal={arXiv preprint arXiv:2308.08155},
  year={2023}
}

@article{xu2026devil,
  title={The Devil Is in the Interface: Evaluating How Tool Architecture Shapes Coding Agent Behavior},
  author={Xu, Xiangzhe and Saghir, Hamidreza and Wu, Qianhui and C{\^o}t{\'e}, Marc-Alexandre and Wang, Tong and Lakkaraju, Kiran and Pei, Kexin and Zhang, Xiangyu},
  journal={arXiv preprint arXiv:2608.11386},
  year={2026}
}

@misc{anthropic2024mcp,
  author       = {{Anthropic}},
  title        = {Introducing the Model Context Protocol},
  year         = {2024},
  month        = nov,
  day          = {25},
  url          = {https://www.anthropic.com/news/model-context-protocol},
  note         = {Accessed: 2026-07-15}
}

@misc{aws2026governing,
  author       = {{AWS}},
  title        = {Governing {AI} Assets at Scale with {MCP} Gateway and Registry},
  year         = {2026},
  month        = jun,
  url          = {https://aws.amazon.com/blogs/opensource/governing-ai-assets-at-scale-with-mcp-gateway-and-registry/},
  note         = {Accessed: 2026-07-15}
}

@misc{arcade2026mcp,
  author       = {{Arcade}},
  title        = {The {MCP} Gateway Pattern: Scaling Agentic Integrations Without Tool Sprawl},
  year         = {2026},
  month        = jan,
  url          = {https://www.arcade.dev/blog/mcp-gateway-pattern/},
  note         = {Accessed: 2026-07-15}
}

@misc{openai2026harness,
  author       = {{OpenAI}},
  title        = {Harness Engineering: Leveraging {Codex} in an Agent-First World},
  year         = {2026},
  month        = feb,
  url          = {https://openai.com/index/harness-engineering/},
  note         = {Accessed: 2026-07-15}
}

@misc{an2025harnesses,
  author       = {{Anthropic}},
  title        = {Effective Harnesses for Long-Running Agents},
  year         = {2025},
  month        = nov,
  url          = {https://anthropic.com/engineering/effective-harnesses-for-long-running-agents},
  note         = {Accessed: 2026-07-15}
}

@misc{langchain2026improving,
  author       = {{LangChain}},
  title        = {Improving {Deep} {Agents} with Harness Engineering},
  year         = {2026},
  month        = feb,
  url          = {https://www.langchain.com/blog/improving-deep-agents-with-harness-engineering},
  note         = {Accessed: 2026-07-15}
}

@misc{openaiagentssdk,
  author       = {{OpenAI}},
  title        = {{OpenAI} {Agents} {SDK}},
  year         = {2025},
  url = {https://github.com/openai/openai-agents-python},
  note         = {Accessed: 2026-07-15}
}

@misc{openclaw2026,
  author       = {{OpenClaw}},
  title        = {Introducing {OpenClaw}},
  year         = {2026},
  url = {https://openclaw.ai/blog/introducing-openclaw},
  note         = {Accessed: 2026-07-15}
}

@misc{arcade,
  author       = {{Arcade AI}},
  title        = {{Arcade}},
  year         = {2026},
  url = {https://www.arcade.dev/},
}

@misc{docker,
  author       = {{Docker}},
  title        = {Docker {MCP} Registry},
  year         = {2025},
  url = {https://github.com/docker/mcp-registry},
}

@misc{microsoftmcpgateway,
  author = {{Microsoft}},
  title  = {{MCP Gateway}},
  url    = {https://github.com/microsoft/mcp-gateway},
  year         = {2025},
}

@misc{unla,
  author       = {{AmoyLab}},
  title        = {Unla},
  year         = {2025},
  url          = {https://github.com/AmoyLab/Unla},
}

@misc{agentgateway,
  author       = {{Agent Gateway}},
  title        = {{AgentGateway}},
  url = {https://github.com/agentgateway/agentgateway},
  year = {2025}
}

@misc{litellm,
  author       = {{BerriAI}},
  title        = {{LiteLLM}},
  url = {https://github.com/BerriAI/litellm},
  year = {2023}
}

@misc{lunarmcpx,
  author       = {{Lunar.dev}},
  title        = {{MCPX}},
  year         = {2025},
  url          = {https://docs.lunar.dev/mcpx/},
}

@misc{awsbedrockgateway,
  author       = {{AWS}},
  title        = {Amazon {Bedrock} {AgentCore} {Gateway}},
  url          = {https://docs.aws.amazon.com/bedrock-agentcore/latest/devguide/gateway.html},
  year = {2025}
}

@misc{enkryptai,
  author       = {{Enkrypt AI}},
  title        = {Secure {MCP} {Gateway}},
  url = {https://github.com/enkryptai/secure-mcp-gateway},
  year = {2025}
}

@misc{klavisstrata,
  author       = {{Klavis AI}},
  title        = {Strata},
  year         = {2025},
  url          = {https://www.klavis.ai/docs/concepts/strata},
}

@misc{bifrost,
  author       = {{Maxim}},
  title        = {{Bifrost} {AI} {Gateway}},
  year         = {2025},
  url = {https://github.com/maximhq/bifrost},
}
\bibliographystyle{iclr2027_conference}

\appendix

\newpage

\section{Additional Results}

\subsection{Real-World Tool Scaling on MCP-Universe}
\label{app:mcp-universe}

This experiment was run in an early, exploratory stage of the project, so it covers only three representative gateways and a smaller range of noisy tools than our main experiments.
We include it to confirm that our tool-scaling trends also hold when the task tools come from real MCP servers, not only in the simulated task environments of \dtap (Section~\ref{sec:task-curation}).
Table~\ref{tab:mcp-universe} reports benign task success for three representative gateways under the OpenAI Agents SDK with GPT-5.1, as the real-world noisy-tool pool grows from $158$ to $358$ tools.

\begin{table}[htbp]
\centering

\caption{Real-world tool scaling on MCP-Universe (OpenAI Agents SDK, GPT-5.1) for three representative gateways. TSR is the fraction of tasks that pass all execution-based checkpoints, and Score is the average fraction of per-task checkpoints passed (partial credit); higher is better for both.}
\footnotesize
\setlength{\tabcolsep}{5pt}
\begin{tabular}{l cc cc cc}
\toprule
\multirow{2}{*}{\textbf{Gateway}}
& \multicolumn{2}{c}{\textbf{158 Noisy Tools}}
& \multicolumn{2}{c}{\textbf{254 Noisy Tools}}
& \multicolumn{2}{c}{\textbf{358 Noisy Tools}} \\
\cmidrule(lr){2-3}
\cmidrule(lr){4-5}
\cmidrule(lr){6-7}
& ~\textbf{Score} & \textbf{TSR (\%)}
& ~\textbf{Score} & \textbf{TSR (\%)}
& ~\textbf{Score} & \textbf{TSR (\%)} \\
\midrule
agentgateway
& 0.73 & 52.3
& 0.66 & 52.0
& 0.75 & 54.0 \\

enkrypt
& 0.60 & 34.7
& 0.62 & 36.0
& 0.52 & 23.3 \\

strata
& 0.68 & 44.7
& 0.68 & 46.0
& 0.68 & 43.7 \\
\bottomrule
\end{tabular}
\label{tab:mcp-universe}
\end{table}

Within this range, benign task success stays largely flat for all three gateways.
Strata, a hierarchical gateway, holds steady ($44.7\%$ to $43.7\%$), which is expected and consistent with its behavior on \dtap: because a hierarchical gateway exposes tools only through discovery, additional noisy tools never enter the agent's context.
Agentgateway, although flattened, is similarly unaffected ($52.3\%$ to $54.0\%$). This also matches \dtap, where agentgateway is the least noise-sensitive of the flattened gateways (Table~\ref{tab:benign-scale}).
We read this as the combination of two factors: the pool here reaches only $358$ tools, well short of the scale at which flattened gateways begin to lose task success in the main experiment, and agentgateway is a comparatively robust flattened implementation.
The only visible decline is enkrypt at the largest pool ($36.0\%$ to $23.3\%$); given the single-run setup and enkrypt's weak-model capability strain under GPT-5.1, we do not over-interpret this point.

\subsection{Indirect Attack Success by Injection Type}
\label{app:asr-channels}
\begin{figure}[htbp]
    \centering
    \begin{tikzpicture}
\begin{groupplot}[
  group style={group size=3 by 2, horizontal sep=1.1cm, vertical sep=1.1cm},
  width=0.35\linewidth, height=3.9cm,
  xtick={0,500,1000,1500}, ymin=0, ymax=80,
  axis line style={line width=0.5pt, draw=black!40},
  every tick/.append style={black!40, line width=0.4pt}, major tick length=2pt,
  tick label style={font=\sffamily\bfseries\tiny}, label style={font=\sffamily\bfseries\scriptsize},
  title style={font=\sffamily\bfseries\scriptsize}, grid=both, grid style={line width=.1pt, draw=gray!25},
  every axis plot/.append style={line width=0.9pt, mark size=1.4pt},
]
\nextgroupplot[title={OpenAI SDK (GPT-5.1)}, ylabel={Env-injection ASR (\%)}, legend entries={agentcore, agentgateway, litellm, lunar, enkrypt, strata, bifrost}, legend to name=asrlegend, legend columns=7, legend style={font=\scriptsize, draw=gray!40, /tikz/every even column/.append style={column sep=6pt}}]
\addplot[color={rgb,255:red,31;green,119;blue,180}, mark=*, densely dashed] coordinates {(0,33.333) (500,25.000) (1000,26.531) (1500,22.917)};
\addplot[color={rgb,255:red,26;green,188;blue,156}, mark=*, densely dashed] coordinates {(0,41.667) (500,29.167) (1000,28.000) (1500,24.490)};
\addplot[color={rgb,255:red,23;green,190;blue,207}, mark=*, densely dashed] coordinates {(0,41.667) (500,16.667) (1000,22.642) (1500,11.765)};
\addplot[color={rgb,255:red,44;green,160;blue,44}, mark=*, densely dashed] coordinates {(0,27.083) (500,31.373) (1000,25.000) (1500,22.642)};
\addplot[color={rgb,255:red,148;green,103;blue,189}, mark=square*, solid] coordinates {(0,4.167) (500,6.250) (1000,4.167) (1500,8.333)};
\addplot[color={rgb,255:red,214;green,39;blue,40}, mark=square*, solid] coordinates {(0,29.167) (500,25.000) (1000,25.000) (1500,18.750)};
\addplot[color={rgb,255:red,255;green,127;blue,14}, mark=square*, solid] coordinates {(0,12.500) (500,12.500) (1000,14.583) (1500,14.583)};
\nextgroupplot[title={OpenAI SDK (GPT-5.4)}]
\addplot[color={rgb,255:red,31;green,119;blue,180}, mark=*, densely dashed] coordinates {(0,35.417) (500,39.583) (1000,37.500) (1500,29.167)};
\addplot[color={rgb,255:red,26;green,188;blue,156}, mark=*, densely dashed] coordinates {(0,29.167) (500,29.167) (1000,33.333) (1500,33.333)};
\addplot[color={rgb,255:red,23;green,190;blue,207}, mark=*, densely dashed] coordinates {(0,31.250) (500,33.333) (1000,29.167) (1500,22.917)};
\addplot[color={rgb,255:red,44;green,160;blue,44}, mark=*, densely dashed] coordinates {(0,35.417) (500,27.083) (1000,31.250) (1500,27.083)};
\addplot[color={rgb,255:red,148;green,103;blue,189}, mark=square*, solid] coordinates {(0,31.250) (500,29.167) (1000,27.083) (1500,33.333)};
\addplot[color={rgb,255:red,214;green,39;blue,40}, mark=square*, solid] coordinates {(0,31.250) (500,27.083) (1000,29.167) (1500,33.333)};
\addplot[color={rgb,255:red,255;green,127;blue,14}, mark=square*, solid] coordinates {(0,33.333) (500,33.333) (1000,33.333) (1500,25.000)};
\nextgroupplot[title={OpenClaw (GPT-5.1)}]
\addplot[color={rgb,255:red,31;green,119;blue,180}, mark=*, densely dashed] coordinates {(0,27.083) (500,27.083) (1000,25.000) (1500,16.327)};
\addplot[color={rgb,255:red,26;green,188;blue,156}, mark=*, densely dashed] coordinates {(0,22.917) (500,22.917) (1000,27.083) (1500,14.583)};
\addplot[color={rgb,255:red,23;green,190;blue,207}, mark=*, densely dashed] coordinates {(0,18.750) (500,20.833) (1000,16.667) (1500,17.647)};
\addplot[color={rgb,255:red,44;green,160;blue,44}, mark=*, densely dashed] coordinates {(0,25.000) (500,18.750) (1000,18.750) (1500,24.490)};
\addplot[color={rgb,255:red,148;green,103;blue,189}, mark=square*, solid] coordinates {(0,12.500) (500,18.750) (1000,14.583) (1500,14.583)};
\addplot[color={rgb,255:red,214;green,39;blue,40}, mark=square*, solid] coordinates {(0,20.833) (500,22.917) (1000,20.833) (1500,22.917)};
\addplot[color={rgb,255:red,255;green,127;blue,14}, mark=square*, solid] coordinates {(0,14.286) (500,18.750) (1000,24.000) (1500,22.917)};
\nextgroupplot[ylabel={Tool-injection ASR (\%)}, xlabel={Number of noisy tools}]
\addplot[color={rgb,255:red,31;green,119;blue,180}, mark=*, densely dashed] coordinates {(0,70.588) (500,56.122) (1000,37.624) (1500,33.333)};
\addplot[color={rgb,255:red,26;green,188;blue,156}, mark=*, densely dashed] coordinates {(0,70.588) (500,71.569) (1000,69.000) (1500,61.386)};
\addplot[color={rgb,255:red,23;green,190;blue,207}, mark=*, densely dashed] coordinates {(0,73.529) (500,51.961) (1000,43.299) (1500,28.283)};
\addplot[color={rgb,255:red,44;green,160;blue,44}, mark=*, densely dashed] coordinates {(0,73.529) (500,55.556) (1000,46.809) (1500,37.113)};
\addplot[color={rgb,255:red,148;green,103;blue,189}, mark=square*, solid] coordinates {(0,14.706) (500,17.647) (1000,17.647) (1500,12.745)};
\addplot[color={rgb,255:red,214;green,39;blue,40}, mark=square*, solid] coordinates {(0,33.333) (500,33.333) (1000,30.392) (1500,31.373)};
\addplot[color={rgb,255:red,255;green,127;blue,14}, mark=square*, solid] coordinates {(0,16.667) (500,19.608) (1000,14.706) (1500,20.588)};
\nextgroupplot[xlabel={Number of noisy tools}]
\addplot[color={rgb,255:red,31;green,119;blue,180}, mark=*, densely dashed] coordinates {(0,54.902) (500,50.000) (1000,51.961) (1500,45.098)};
\addplot[color={rgb,255:red,26;green,188;blue,156}, mark=*, densely dashed] coordinates {(0,53.922) (500,51.961) (1000,50.000) (1500,45.098)};
\addplot[color={rgb,255:red,23;green,190;blue,207}, mark=*, densely dashed] coordinates {(0,53.922) (500,50.980) (1000,44.118) (1500,41.176)};
\addplot[color={rgb,255:red,44;green,160;blue,44}, mark=*, densely dashed] coordinates {(0,52.941) (500,52.941) (1000,48.039) (1500,49.020)};
\addplot[color={rgb,255:red,148;green,103;blue,189}, mark=square*, solid] coordinates {(0,41.176) (500,44.118) (1000,38.235) (1500,42.157)};
\addplot[color={rgb,255:red,214;green,39;blue,40}, mark=square*, solid] coordinates {(0,34.314) (500,31.373) (1000,34.314) (1500,32.353)};
\addplot[color={rgb,255:red,255;green,127;blue,14}, mark=square*, solid] coordinates {(0,15.686) (500,21.569) (1000,24.510) (1500,19.608)};
\nextgroupplot[xlabel={Number of noisy tools}]
\addplot[color={rgb,255:red,31;green,119;blue,180}, mark=*, densely dashed] coordinates {(0,50.000) (500,42.157) (1000,40.196) (1500,36.634)};
\addplot[color={rgb,255:red,26;green,188;blue,156}, mark=*, densely dashed] coordinates {(0,50.000) (500,56.863) (1000,52.941) (1500,51.961)};
\addplot[color={rgb,255:red,23;green,190;blue,207}, mark=*, densely dashed] coordinates {(0,50.980) (500,41.176) (1000,36.275) (1500,34.343)};
\addplot[color={rgb,255:red,44;green,160;blue,44}, mark=*, densely dashed] coordinates {(0,54.902) (500,42.157) (1000,37.255) (1500,39.604)};
\addplot[color={rgb,255:red,148;green,103;blue,189}, mark=square*, solid] coordinates {(0,19.608) (500,17.647) (1000,21.569) (1500,15.686)};
\addplot[color={rgb,255:red,214;green,39;blue,40}, mark=square*, solid] coordinates {(0,17.647) (500,18.627) (1000,19.608) (1500,22.549)};
\addplot[color={rgb,255:red,255;green,127;blue,14}, mark=square*, solid] coordinates {(0,8.911) (500,11.765) (1000,10.000) (1500,8.824)};
\end{groupplot}
\node[anchor=north, yshift=-3mm] at (current bounding box.south) {\pgfplotslegendfromname{asrlegend}};
\end{tikzpicture}
    \caption{Indirect attack success rate by injection type as the tool set scales: environment injection (top) and tool injection (bottom), for each agent configuration (columns). One line per gateway; flattened gateways are dashed, hierarchical gateways solid.}
    \label{fig:asr-channels-full}
\end{figure}
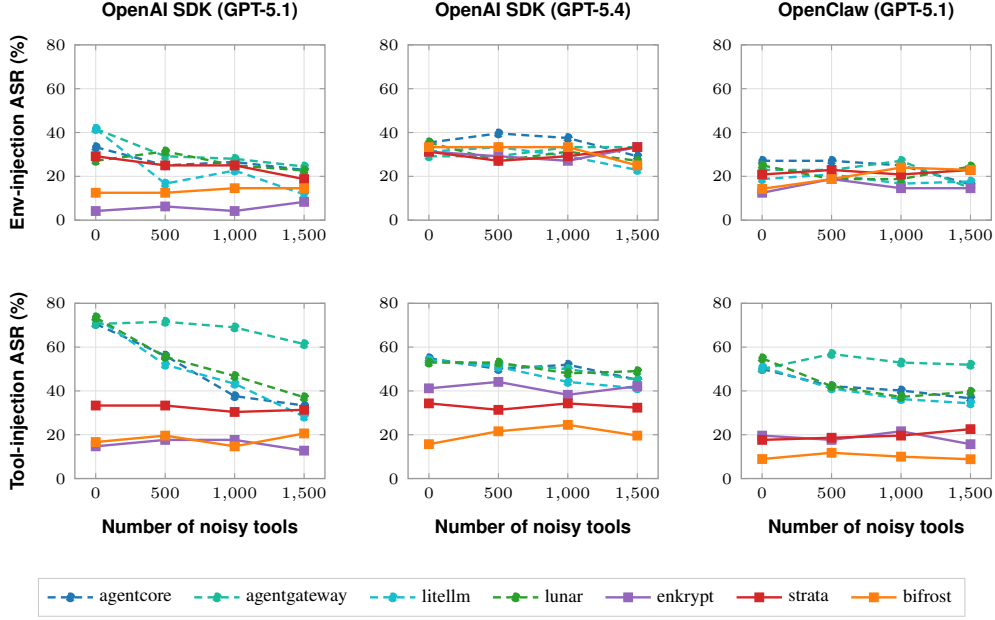

Among the 150 indirect attack tasks in our dataset, 48 are environment injection attack tasks and 102 are tool injection attack tasks.
Figure~\ref{fig:asr-channels-full} gives the full per-gateway decomposition for all three agent configurations.
The tool-injection gap between the two architecture types appears in every configuration, but the two OpenAI Agents SDK settings each add a caveat that the OpenClaw view in the main text removes.

Under GPT-5.4 the two architecture types are capability-matched (mean TSR $85.1\%$ and $87.2\%$), and the same pattern holds: tool injection separates them (the flattened architecture's mean runs from $53.9\%$ down to $45.1\%$ against a hierarchical architecture near $31\%$), whereas environment injection does not (the two means stay within about $3$ points at every noisy-tool count).
Hierarchy alone, however, is not what protects the agent: among the hierarchical gateways, tool-injection ASR runs from roughly $20\%$ (bifrost) through $33\%$ (strata) to $41\%$ (enkrypt), a spread as wide as the gap between the architecture types, so how each interface exposes an injected tool matters as much as whether it is hierarchical.

Under GPT-5.1 with the OpenAI Agents SDK the hierarchical architecture shows the lowest attack success of all (tool injection near $21\%$, environment injection near $15\%$), but this must not be read as protection.
The same configuration collapses hierarchical benign task success to $47.5\%$ (Table~\ref{tab:benign-scale}), so an agent that cannot drive tool discovery also fails to carry out an injected goal; its apparent robustness is capability collapse, not safety.
The wider environment-injection gap here ($35.9\%$ for flattened against $15.3\%$ for hierarchical at $0$ tools) is likewise an effect of the weak model rather than the gateway, since a gateway does not govern what an environment returns.

\subsection{Detailed Efficiency Results}
\label{app:efficiency}

\begin{figure}[htbp]
    \centering
    \begin{tikzpicture}
\begin{groupplot}[
  group style={group size=3 by 2, horizontal sep=1.45cm, vertical sep=1.55cm},
  width=0.32\linewidth, height=3.6cm,
  xtick={0,500,1000,1500},
  axis line style={line width=0.5pt, draw=black!40},
  every tick/.append style={black!40, line width=0.4pt}, major tick length=2pt,
  tick label style={font=\sffamily\bfseries\tiny},
  label style={font=\sffamily\bfseries\scriptsize},
  title style={font=\sffamily\bfseries\scriptsize},
  xticklabel style={font=\sffamily\bfseries\tiny}, yticklabel style={font=\sffamily\bfseries\tiny},
  grid=both, grid style={line width=.1pt, draw=gray!25},
  every axis plot/.append style={line width=0.9pt, mark size=1.1pt},
]
\nextgroupplot[ylabel={\sffamily\bfseries\scriptsize Task time (s)}, ymin=0, legend entries={agentcore, agentgateway, litellm, lunar, enkrypt, strata, bifrost}, legend to name=costlegendapp, legend columns=-1, legend style={font=\scriptsize, draw=gray!40, /tikz/every even column/.append style={column sep=6pt}}]
\addplot[color={rgb,255:red,31;green,119;blue,180}, mark=*, densely dashed] coordinates {(0,21.472) (500,35.041) (1000,60.082) (1500,74.075)};
\addplot[color={rgb,255:red,26;green,188;blue,156}, mark=*, densely dashed] coordinates {(0,30.003) (500,46.514) (1000,58.714) (1500,79.988)};
\addplot[color={rgb,255:red,23;green,190;blue,207}, mark=*, densely dashed] coordinates {(0,27.786) (500,45.734) (1000,62.669) (1500,77.926)};
\addplot[color={rgb,255:red,44;green,160;blue,44}, mark=*, densely dashed] coordinates {(0,36.832) (500,42.331) (1000,59.155) (1500,80.181)};
\addplot[color={rgb,255:red,148;green,103;blue,189}, mark=square*, solid] coordinates {(0,27.331) (500,38.257) (1000,37.886) (1500,44.943)};
\addplot[color={rgb,255:red,214;green,39;blue,40}, mark=square*, solid] coordinates {(0,25.202) (500,27.287) (1000,25.214) (1500,25.814)};
\addplot[color={rgb,255:red,255;green,127;blue,14}, mark=square*, solid] coordinates {(0,39.992) (500,38.751) (1000,39.704) (1500,41.965)};
\nextgroupplot[ylabel={\sffamily\bfseries\scriptsize Tool-call latency (s)}, ymin=0]
\addplot[color={rgb,255:red,31;green,119;blue,180}, mark=*, densely dashed] coordinates {(0,1.273) (500,1.363) (1000,1.849) (1500,1.394)};
\addplot[color={rgb,255:red,26;green,188;blue,156}, mark=*, densely dashed] coordinates {(0,0.272) (500,0.285) (1000,0.282) (1500,0.276)};
\addplot[color={rgb,255:red,23;green,190;blue,207}, mark=*, densely dashed] coordinates {(0,0.507) (500,0.483) (1000,0.539) (1500,0.532)};
\addplot[color={rgb,255:red,44;green,160;blue,44}, mark=*, densely dashed] coordinates {(0,0.288) (500,0.313) (1000,0.294) (1500,0.303)};
\addplot[color={rgb,255:red,148;green,103;blue,189}, mark=square*, solid] coordinates {(0,1.712) (500,2.708) (1000,3.612) (1500,3.466)};
\addplot[color={rgb,255:red,214;green,39;blue,40}, mark=square*, solid] coordinates {(0,0.195) (500,0.183) (1000,0.175) (1500,0.189)};
\addplot[color={rgb,255:red,255;green,127;blue,14}, mark=square*, solid] coordinates {(0,0.105) (500,0.103) (1000,0.103) (1500,0.106)};
\nextgroupplot[ylabel={\sffamily\bfseries\scriptsize Tokens/turn (k)}, ymin=0]
\addplot[color={rgb,255:red,31;green,119;blue,180}, mark=*, densely dashed] coordinates {(0,8.628) (500,63.140) (1000,143.748) (1500,216.131)};
\addplot[color={rgb,255:red,26;green,188;blue,156}, mark=*, densely dashed] coordinates {(0,8.174) (500,63.695) (1000,150.147) (1500,238.713)};
\addplot[color={rgb,255:red,23;green,190;blue,207}, mark=*, densely dashed] coordinates {(0,8.234) (500,63.761) (1000,150.966) (1500,239.153)};
\addplot[color={rgb,255:red,44;green,160;blue,44}, mark=*, densely dashed] coordinates {(0,8.384) (500,64.199) (1000,150.953) (1500,238.899)};
\addplot[color={rgb,255:red,148;green,103;blue,189}, mark=square*, solid] coordinates {(0,7.415) (500,6.820) (1000,8.293) (1500,9.327)};
\addplot[color={rgb,255:red,214;green,39;blue,40}, mark=square*, solid] coordinates {(0,4.130) (500,4.117) (1000,4.754) (1500,5.419)};
\addplot[color={rgb,255:red,255;green,127;blue,14}, mark=square*, solid] coordinates {(0,8.645) (500,8.001) (1000,8.802) (1500,8.598)};
\nextgroupplot[ylabel={\sffamily\bfseries\scriptsize Task time (s)}, ymin=0, xlabel={\sffamily\bfseries\scriptsize Number of noisy tools}]
\addplot[color={rgb,255:red,31;green,119;blue,180}, mark=*, densely dashed] coordinates {(0,19.668) (500,29.825) (1000,44.135) (1500,61.132)};
\addplot[color={rgb,255:red,26;green,188;blue,156}, mark=*, densely dashed] coordinates {(0,19.085) (500,31.895) (1000,44.971) (1500,65.908)};
\addplot[color={rgb,255:red,23;green,190;blue,207}, mark=*, densely dashed] coordinates {(0,21.965) (500,33.739) (1000,45.589) (1500,64.758)};
\addplot[color={rgb,255:red,44;green,160;blue,44}, mark=*, densely dashed] coordinates {(0,22.634) (500,37.360) (1000,47.528) (1500,64.596)};
\addplot[color={rgb,255:red,148;green,103;blue,189}, mark=square*, solid] coordinates {(0,34.516) (500,54.442) (1000,56.349) (1500,58.345)};
\addplot[color={rgb,255:red,214;green,39;blue,40}, mark=square*, solid] coordinates {(0,24.627) (500,24.557) (1000,23.746) (1500,27.799)};
\addplot[color={rgb,255:red,255;green,127;blue,14}, mark=square*, solid] coordinates {(0,31.123) (500,31.362) (1000,34.118) (1500,30.057)};
\nextgroupplot[ylabel={\sffamily\bfseries\scriptsize Tool-call latency (s)}, ymin=0, xlabel={\sffamily\bfseries\scriptsize Number of noisy tools}]
\addplot[color={rgb,255:red,31;green,119;blue,180}, mark=*, densely dashed] coordinates {(0,1.249) (500,1.283) (1000,1.281) (1500,1.319)};
\addplot[color={rgb,255:red,26;green,188;blue,156}, mark=*, densely dashed] coordinates {(0,0.267) (500,0.286) (1000,0.289) (1500,0.272)};
\addplot[color={rgb,255:red,23;green,190;blue,207}, mark=*, densely dashed] coordinates {(0,0.508) (500,0.501) (1000,0.503) (1500,0.475)};
\addplot[color={rgb,255:red,44;green,160;blue,44}, mark=*, densely dashed] coordinates {(0,0.287) (500,0.287) (1000,0.280) (1500,0.288)};
\addplot[color={rgb,255:red,148;green,103;blue,189}, mark=square*, solid] coordinates {(0,2.574) (500,5.079) (1000,5.244) (1500,5.356)};
\addplot[color={rgb,255:red,214;green,39;blue,40}, mark=square*, solid] coordinates {(0,0.223) (500,0.216) (1000,0.211) (1500,0.222)};
\addplot[color={rgb,255:red,255;green,127;blue,14}, mark=square*, solid] coordinates {(0,0.146) (500,0.131) (1000,0.209) (1500,0.135)};
\nextgroupplot[ylabel={\sffamily\bfseries\scriptsize Tokens/turn (k)}, ymin=0, xlabel={\sffamily\bfseries\scriptsize Number of noisy tools}]
\addplot[color={rgb,255:red,31;green,119;blue,180}, mark=*, densely dashed] coordinates {(0,8.273) (500,63.238) (1000,143.936) (1500,215.939)};
\addplot[color={rgb,255:red,26;green,188;blue,156}, mark=*, densely dashed] coordinates {(0,8.268) (500,63.505) (1000,150.056) (1500,238.517)};
\addplot[color={rgb,255:red,23;green,190;blue,207}, mark=*, densely dashed] coordinates {(0,8.309) (500,63.765) (1000,150.339) (1500,238.796)};
\addplot[color={rgb,255:red,44;green,160;blue,44}, mark=*, densely dashed] coordinates {(0,8.375) (500,64.058) (1000,151.040) (1500,239.941)};
\addplot[color={rgb,255:red,148;green,103;blue,189}, mark=square*, solid] coordinates {(0,14.982) (500,14.821) (1000,18.923) (1500,22.785)};
\addplot[color={rgb,255:red,214;green,39;blue,40}, mark=square*, solid] coordinates {(0,3.538) (500,3.964) (1000,4.625) (1500,5.412)};
\addplot[color={rgb,255:red,255;green,127;blue,14}, mark=square*, solid] coordinates {(0,8.748) (500,9.128) (1000,9.407) (1500,9.745)};
\end{groupplot}
\node[anchor=south, yshift=2mm, font=\sffamily\bfseries\scriptsize] at ($(group c1r1.north)!0.5!(group c3r1.north)$) {OpenAI SDK (GPT-5.1)};
\node[anchor=south, yshift=2mm, font=\sffamily\bfseries\scriptsize] at ($(group c1r2.north)!0.5!(group c3r2.north)$) {OpenAI SDK (GPT-5.4)};
\node[anchor=north, yshift=-3mm] at (current bounding box.south) {\pgfplotslegendfromname{costlegendapp}};
\end{tikzpicture}
    \caption{Interaction cost as the tool set scales, under the OpenAI Agents SDK with GPT-5.1 (top) and GPT-5.4 (bottom): task time (left), average tool-call latency (middle), and tokens per turn (right). Flattened gateways are dashed, hierarchical gateways solid.}
    \label{fig:cost-appendix}
\end{figure}
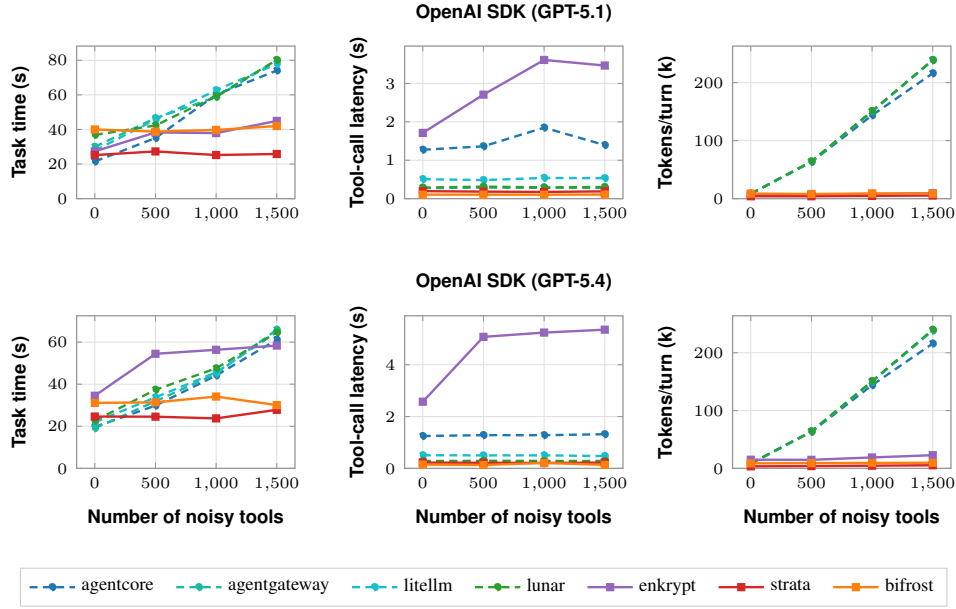

Figure~\ref{fig:cost-appendix} extends the interaction-cost analysis of Section~\ref{sec:results} to both OpenAI Agents SDK models and adds tool-call latency as a third metric.
The task-time and token trends match the main text: flattened gateways grow roughly linearly with the tool set in both, whereas hierarchical gateways stay nearly flat.
The two token curves are almost identical across the models, confirming that per-turn token cost is set by the number of injected noisy tools rather than the model.

Tool-call latency exposes a difference that the aggregate task time hides.
For flattened gateways the average tool call is cheap and roughly constant (about $0.6$s) irrespective of the number of injected noisy tools, since the gateway simply forwards the call to the upstream server.
Hierarchical gateways split sharply: strata and bifrost are even faster per call (around $0.2$s and $0.1$s), whereas enkrypt is far slower and grows with the pool (from $2.6$s to $5.4$s under GPT-5.4), because its layered discovery meta-tools return large server-and-tool listings on every call (Section~\ref{sec:gateway}).
Enkrypt therefore trades many cheap turns for a few expensive discovery calls, which accounts for much of its elevated task time.

\Needspace*{14\baselineskip}
\subsection{Detailed MCP Gateway Analysis}
\label{app:implementation}

\begin{wraptable}{r}{0.4\textwidth}
  \centering
  \caption{Number of task runs in which the AgentCore search tool was invoked. Each agent includes 1,800 task runs.}
  \label{tab:appendix_results}
  \footnotesize
  \begin{tabular}{lc}
    \toprule
    Agent & Search Used \\
    \midrule
    OpenAI SDK (GPT-5.1)          & 11  \\
    OpenAI SDK (GPT-5.4)          & 19  \\
    OpenClaw (GPT-5.1)    & 1   \\
    \bottomrule
  \end{tabular}
\end{wraptable}

\paragraph{AgentCore Search Tool Effectiveness.}
Among the flattened gateways evaluated in this work, AgentCore is unique in providing a built-in search tool that semantically matches tools to the agent's intent. To examine whether this additional search tool offers any practical benefit, we count the number of task runs in which the agent actually invoked the search tool. As shown in Table~\ref{tab:appendix_results}, in 1,800 runs per agent setting, search tool usage is negligible.
We speculate that the agent directly locates the tools relevant to the task and proceeds without invoking search, since all necessary tools are already visible.
Besides, LLM agents are trained to minimize unnecessary tool calls.
These results suggest that a search tool provides limited utility when the full tool list is already exposed.

\paragraph{Enkrypt.}
Enkrypt adopts a hierarchical mechanism, which naturally yields good token efficiency.
However, its actual implementation severely undermines this benefit. Specifically, Enkrypt returns tool definitions as JSON-encoded configuration blobs that include extensive metadata irrelevant to the agent, such as internal runtime commands, guardrail settings, and connection parameters. Furthermore, its default tool entrypoint attempts to return the full tool list at once but frequently fails due to timeouts when the available tool set is large.
When the underlying model is weak, the agent often attempts to skip discovery steps and directly invokes tools with hallucinated names and parameters, which leads to poor task success rates.

\paragraph{Bifrost.}
Bifrost provides 4 meta tools to the agent: \texttt{executeToolCode}, \texttt{getToolDocs}, \texttt{listToolFiles}, and \texttt{readToolFile}. These meta tools creates a virtual filesystem where each server is represented as a Python stub file(.pyi). To reduce the token cost, the detailed description of the parameters are truncated in the file and agent need to call \texttt{getToolDocs} to get the full details. For weak models like GPT-5.1, the generated code often mishandles special characters such as newline characters inside string literals, writing as a physical line break rather than the escaped sequence (e.g. \texttt{split('\textbackslash n')}). This produces incorrect Python code and is consequently rejected by the parser, making the gateway effectively unusable for those models.
For indirect malicious tasks, the truncated tool description protects the agent from seeing the malicious message and thus lowers the ASR.

\paragraph{Strata.}
Strata provides 5 meta tools to the agent: \texttt{discover\_server\_actions},\\
\texttt{get\_action\_details}, \texttt{execute\_action}, \texttt{search\_documentation},\\
and \texttt{handle\_auth\_failure}. The agent must first search the required tools using a semantic query and get the detail of the tool before invoking. However, when the model is not well trained for this situation like GPT-5.1, it often skips the discovery process and directly invokes hallucinated tools or parameters.

\end{document}